\documentclass[10pt,journal,compsoc]{IEEEtran}

\ifCLASSOPTIONcompsoc
  \usepackage[nocompress]{cite}
  \usepackage{multicol}
    \usepackage{multirow}
    \usepackage{color}
    \usepackage{xspace}
    \usepackage{times}
    \usepackage{epsfig}
    \usepackage{epstopdf}
    \usepackage{graphicx}
    \usepackage{amssymb}
    \usepackage{amsmath}
    \usepackage{url}
    \usepackage{multicol}
    \usepackage{breqn}
    \usepackage{booktabs}
    \usepackage[table,xcdraw]{xcolor}
    \usepackage{tabularx}
    \usepackage{multirow}
    \usepackage{xcolor,colortbl}
    \usepackage{cite}
    \definecolor{Gray}{gray}{0.85}
    \newcolumntype{a}{>{\columncolor{Gray}}c}

    \usepackage{amsfonts}
    \usepackage{bbm}
\else
  \usepackage{cite}
\fi
\newcommand{\wx}[1]{\textcolor{black}{#1}}
\newcommand{\wxr}[1]{\textcolor{black}{#1}}

\newcommand{\lqq}[1]{\textcolor{black}{#1}}

\ifCLASSINFOpdf
\else
\fi
\begin{document}
%
\title{Monocular BEV Perception of Road Scenes via Front-to-Top View Projection}
%
%
%
%

\author{\normalsize{Wenxi Liu, Qi Li, Weixiang Yang, Jiaxin Cai, Yuanlong Yu, Yuexin Ma, Shengfeng He, Jia Pan}}

\markboth{Journal of \LaTeX\ Class Files,~Vol.~14, No.~8, August~2015}%
{Shell \MakeLowercase{\textit{et al.}}: Bare Advanced Demo of IEEEtran.cls for IEEE Computer Society Journals}
%



\IEEEtitleabstractindextext{%
\begin{abstract}
HD map reconstruction is crucial for autonomous driving. LiDAR-based methods are limited due to expensive sensors and time-consuming computation. Camera-based methods usually need to perform road segmentation and view transformation separately, which often causes distortion and missing content. 
To push the limits of the technology, we present a novel framework that reconstructs a local map formed by road layout and vehicle occupancy in the bird's-eye view given a front-view monocular image only. 
We propose a front-to-top view projection (FTVP) module, which takes the constraint of cycle consistency between views into account and makes full use of their correlation to strengthen the view transformation and scene understanding. 
In addition, we also apply multi-scale FTVP modules to propagate the rich spatial information of low-level features to mitigate spatial deviation of the predicted object location.
Experiments on public benchmarks show that our method achieves the state-of-the-art performance in the tasks of road layout estimation, vehicle occupancy estimation, and multi-class semantic estimation. For multi-class semantic estimation, in particular, our model outperforms all competitors by a large margin.
Furthermore, our model runs at 25 FPS on a single GPU, which is efficient and applicable for real-time panorama HD map reconstruction.
\end{abstract}

}

\maketitle

\IEEEdisplaynontitleabstractindextext

%
\IEEEpeerreviewmaketitle

\ifCLASSOPTIONcompsoc
\IEEEraisesectionheading{\section{Introduction}\label{sec:introduction}}
\else
\section{Introduction}
\label{sec:introduction}
\fi

\IEEEPARstart{W}{ith} the rapid progress of autonomous driving technologies, there have been many recent efforts on related research topics, e.g., scene layout estimation~\cite{Geiger14,sun2019leveraging,roddick2020predicting,liu2020understanding,liang2019convolutional,2018Learning,2019A}, 3D object detection~\cite{2016Monocular,srivastava2019learning,roddick2018orthographic,simonelli2019disentangling,ding2020learning,2020GS3D}, vehicle behavior prediction~\cite{mozaffari2020deep,hong2019rules,kim2020advisable,ma2019trafficpredict}, and lane detection~\cite{hou2019learning,zou2019robust,philion2019fastdraw}. 

\begin{figure}
    \centering
    \includegraphics[trim=1.3cm 0 0 0,clip,width=0.5\textwidth]{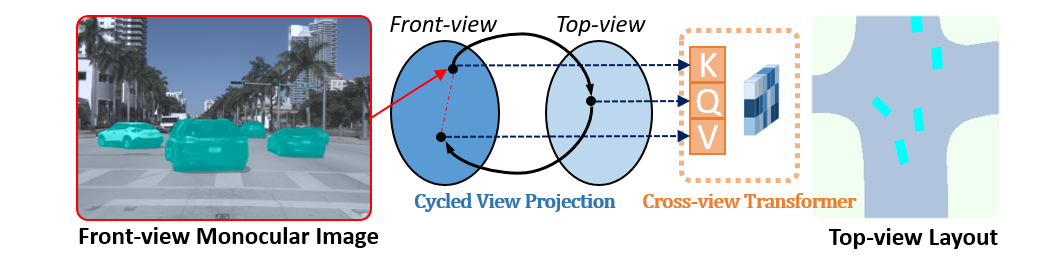}
    \caption{Given a front-view monocular image, we propose a front-to-top view projection module consisting of a cycle structure that bridges the features of the front-view and top-view in their respective domains, as well as a cross-view transformer that correlates views attentively to facilitate the road layout estimation.}
    \label{fig:teaser}
\end{figure}

Among these tasks, high-definition map (HD map) reconstruction is fundamental and critical for perception, prediction, and planning of autonomous driving. 
Its major issues are concerned with the estimation of a local map including the road layout as well as the occupancies of nearby vehicles in the 3D world. Existing techniques rely on expensive sensors like LiDAR and require time-consuming computation for cloud point data. In addition, camera-based techniques usually need to perform road segmentation and view transformation separately, which causes distortion and the absence of content.
To push the limits of the technology, our work aims to address the realistic yet challenging problem of estimating the road layout and vehicle occupancy in top-view, or bird's-eye view (BEV), given a single monocular front-view image (see Fig.~\ref{fig:teaser}).
However, due to the large view gap and severe view deformation, understanding and estimating the top-view scene layout from the front-view image is an extremely difficult problem, even for a human observer. 
The same scene has significantly different appearances in the images of bird's-eye-view and the front-view. Thus, parsing and projecting the road scenes of front-view to top-view require the ability to fully exploit the information of the front-view image and innate reasoning about the unseen regions. 

Traditional methods (e.g.,~\cite{lin2012vision,tseng2013image}) focus on investigating the perspective transformation by estimating the camera parameters and performing image coordinate transformation, but gaps in the resulting BEV feature maps caused by geometric warping lead to poor results.
Recent deep learning-based approaches \cite{zhu2018generative,regmi2018cross} mainly rely on the hallucination capability of deep Convolutional Neural Networks (CNNs) to infer the unseen regions between views. 
In general, instead of modeling the correlation between views, these methods directly leverage CNNs to learn the view projection models in a supervised manner.
These models require deep network structures to propagate and transform the features of the front-view through multiple layers to spatially align with the top-view layout. However, due to the locally confined receptive fields of convolutional layers, it is difficult to fit a view projection model and identify the vehicles of small scales.



To address these concerns, we derive a novel framework to estimate the road layout and vehicle occupancies from the top-view given a single monocular front-view image.
To handle the large discrepancy between views, we present a \wx{Front-to-Top View Projection (FTVP) module} in our network, which is composed of two sub-modules: \textit{Cycled View Projection} (CVP) module bridges the view features in their respective domains and \textit{Cross-View Transformer} (CVT) correlates the views, as shown in Fig.~\ref{fig:teaser}. Specifically, the CVP projects views using a multi-layer perceptron (MLP), which overtakes the standard information flow passing through convolutional layers, and involves the constraint of cycle consistency to retain the features relevant for view projection. { In other words, transforming front-views to top-views requires a global spatial transformation over the visual features. However, standard CNN layers only allow local computation over feature maps, and it takes several layers to obtain a sufficiently large receptive field. On the other hand, fully connected layers can better facilitate the \wx{front-to-top view projection}.} Then CVT explicitly correlates the features of the views before and after projection obtained from CVP, which can significantly enhance the features after view projection. We involve a feature selection scheme in CVT, which leverages the associations of both views to extract the most relevant information.

\wx{However, at the core of our network model, the low-resolution encoded top-view features may result in the spatial deviation of the predicted object location, because of the information loss of heavily compressed features. 
To alleviate this issue, low-level visual features need to be exploited to refine the upsampled features. To bridge the low-level visual features from the front-view and the upsampled features, 
we propose applying multi-scale FTVP modules to project and propagate the low-level features. Concretely, the front-view features from the downsampling stream are first projected and transformed to top-view ones via the FTVP modules. Then all the projected features are concatenated with the corresponding top-view features of the upsampling stream. 
Moreover, to further enhance the representation of top-view features from different scales, deep supervision is employed to supervise the segmentation heads.}

In comprehensive experiments, we demonstrate that our \wx{front-to-top view projection} module can effectively elevate the performance of road layout and vehicle occupancy estimation. For both tasks, we compare our model against the state-of-the-art methods on public benchmarks and demonstrate that our model is superior to all the other methods. 
It is worth noting that, for estimating vehicle occupancies, our model achieves a significant advantage over the competing methods by at least $34.8\%$ in the \textit{KITTI 3D Object} dataset and by at least $46.4\%$ in the \textit{Argoverse} dataset. 
Furthermore, we also validate our framework by predicting semantic segmentation results with more than seven classes in both the \textit{Argoverse} and \textit{NuScenes} datasets. We show that our method surpasses the latest off-the-shelf methods, which reflects the generality and rationality of our proposed model. Last but not least, we show that our framework is able to process $1024\times 1024$ images in $25$ FPS using a single Titan XP GPU, and it is applicable for real-time reconstruction of a panorama HD map. 

Overall, the contributions of our paper are:
\begin{itemize}
    \item We propose a novel framework that reconstructs a local map formed by a top-view road scene layout and vehicle occupancy using only a single monocular front-view image.
    We propose a \wx{front-to-top view projection} module, which leverages the cycle consistency between views and their correlation to strengthen the view transformation.
    \wx{\item We also utilize multi-scale FTVP modules that project and propagate front-view features of scales to refine the representation of top-view features, which provides richer cues for precise estimation of object location.}
    \item On public benchmarks, we demonstrate that our model achieves the state-of-the-art performance for the tasks of \wx{road layout, vehicle occupancy, and multi-class semantic estimation.}
\end{itemize}

Note that a preliminary version of this work was presented as \cite{yang2021projecting}. This submission extends \cite{yang2021projecting} the methodology and experiment in the following aspects. First, we renovate our model by adjusting the structure of the front-to-top view projection module in which we rectify the projected features by utilizing the raw front-view features for supplementing visual information. This improves the expressive ability of the final top-view features and enhances the robustness of the module structure (Sec.~\ref{sec:First}).
Second, our previous model \cite{yang2021projecting} tends to cause the spatial deviation in vehicle occupancy prediction and road layout estimation, due to upsampling the low-resolution features to higher-resolution ones. To address this concern, we re-design the framework by directly projecting multi-scale features from the downsampling stream via the multi-scale FTVP modules to strengthen the upsampling process using the skip connection (Sec.~\ref{sec:Second}). 
\lqq{With the above improvements, our proposed model performs better than preliminary version on all datasets. Especially on the \textit{KITTI 3D Object} dataset, our method achieves 4.9\% and 17.5\% improvements in terms of mIOU and mAP.}
Third, we extend our model to address multi-class segmentation problems with more than seven classes, which shows our model is able to handle scene parsing for diverse objects not limited to vehicles and road (Sec.~\ref{sec:Third}).
\lqq{We visualize the comparison results on multi-class semantic estimation in Fig.~\ref{fig:fig_nu} to embody the advantage of our method over others.} 
Finally, we evaluate and discuss the effectiveness of our proposed framework compared with our preliminary version and conduct additional experiments against the state-of-the-art results by estimating the multi-class segmentation maps in the \textit{Argoverse} and \textit{NuScenes} datasets (Sec.~\ref{sec:Third} and \ref{sec:Fourth}). 
Furthermore, we demonstrate that our proposed network is lightweight, and it can achieve real-time performance on a single Titan XP GPU.

\wx{In the remainder of this paper, we first review the related work in Sec.~\ref{sec:related_works}. We will then describe the overall framework and introduce the front-to-top view projection module in Sec.~\ref{sec:Method}. In Sec.~\ref{sec:Results}, the experimental results are demonstrated and discussed. Finally, we summarize our work in Sec.~\ref{sec:Conclusion}.}

\section{Related Work}
\label{sec:related_works}

\begin{figure*}
    \centering
    \includegraphics[width=0.98\textwidth]{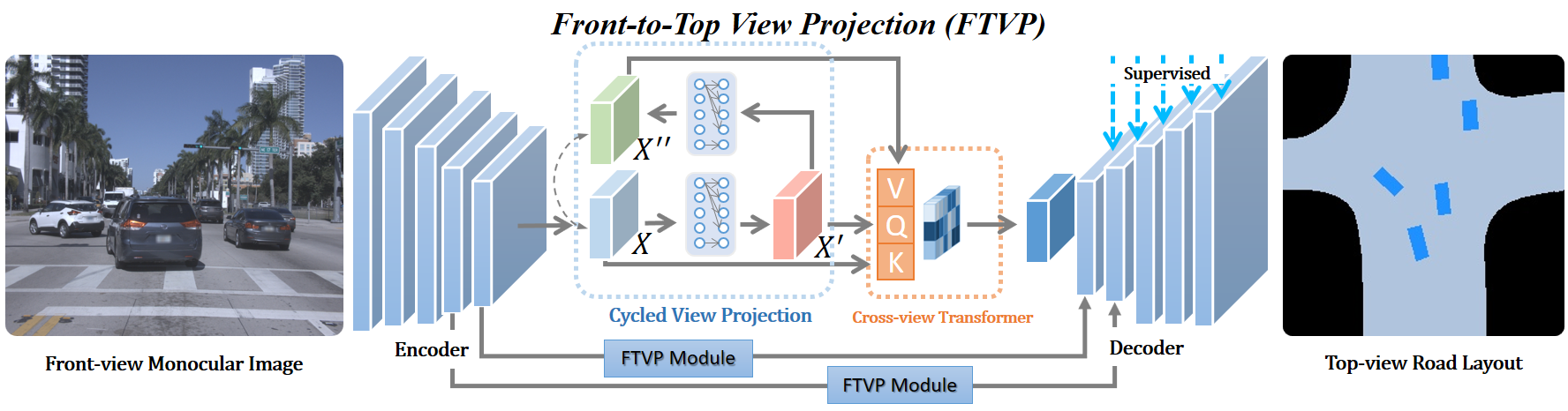}
    \caption{As illustrated, our network aims to transform the front-view monocular image to the top-view road layout. The main component of our proposed \textit{front-to-top view projection module} is the cycled view projection (CVP) and the cross-view transformer (CVT), which projects the features from the front-view domain, $X$, to the top-view domain, $X'$. In CVP, it first utilizes an MLP-based cycle structure to retain the confident features for view projection in $X''$, and then CVT correlates the features of both views to attentively enhance $X'$. \wx{Moreover, we also utilize multi-scale FTVP modules to project the front-view features on varied scales to top-down view, which brings rich spatial information for top-view features in the upsampling stream to alleviate the spatial deviation on object location estimation.}}
    \label{fig:framework}
\end{figure*}

In this section, we survey the related literature on road layout estimation, vehicle detection, and street view synthesis on top-view representation. We also introduce the recent progress of transformers on vision tasks.

\textbf{BEV-based road layout estimation and vehicle detection.} Most road scene parsing works focus on semantic segmentation \cite{fan2020sne,teichmann2018multinet,yang2018denseaspp,yu2018bisenet,fu2019dual}, while there are a few attempts that derive top-view representation for road layout \cite{2012Automatic,2016HD,Geiger14,2017Predicting,2017Cognitive,2018Learning,2019A,lu2019monocular}. 
Among these methods, Schulter et al. \cite{2018Learning} propose estimating an occlusion-reasoned road layout on top-view from a single color image by depth estimation and semantic segmentation.
VED \cite{lu2019monocular} proposes a variational autoencoder (VAE) model to predict road layout from a given image, but without attempting to reason about the unseen layout from observation. VPN \cite{pan2020cross} presents a cross-view semantic segmentation by transforming and fusing the observation from multiple cameras. {~\cite{philion2020lift, roddick2020predicting} directly transform features from images to 3D space and finally to bird’s-eye-view (BEV) grids.}
On the other hand, 
many monocular image-based 3D vehicle detection techniques have been developed (e.g., \cite{2016Monocular,2017Mousavian,2020GS3D,lang2019pointpillars}). Several methods handle this problem by mapping the monocular image to the top-view. For instance, \cite{roddick2018orthographic} proposes mapping a monocular image to the top-view representation and treats 3D object detection as a task of 2D segmentation. \wx{2D-Lift~\cite{dwivedi2021bird} proposes the BEV feature transform layer to transform 2D image features to the BEV space, which exploits depth maps and 3D point cloud.} BirdGAN \cite{srivastava2019learning} also leverages adversarial learning for mapping images to bird's-eye-view. \wx{STA \cite{saha2021enabling} and Stitch \cite{can2022understanding} aggregate the temporal information to produce the final segmentation results through the transformation module.}
As another related work, \cite{wang2019monocular} does not focus on explicit scene layout estimation, focusing instead on the motion planning side. 
Most related to our work, \cite{2020MonoLayout} presents a unified model to tackle the task of road layout (static scene) and traffic participant (dynamic scene) estimations from a single image. In contrast, we propose an approach to explicitly model the large view projection that learns the spatial information to produce high-quality results.


\textbf{View transformation and synthesis.} 
Traditional methods (e.g.~\cite{lin2012vision,tseng2013image,huang2018lane}) have been proposed to handle the perspective transformation in traffic scenes. 
With the progress of deep learning-based methods, \cite{zhu2018generative} proposes a pioneering work to generate the bird's-eye-view based on the driver's view. They treat cross-view synthesis as an image translation task and adopt a GAN-based framework to accomplish it. Due to the difficulty in collecting annotation for real data, their model is trained from video game data. {\cite{abbas2019geometric} focuses exclusively on warping camera images to BEV images without performing any downstream tasks such as object detection.} 
Recent attempts~\cite{regmi2018cross,tang2019multi} on view synthesis aim to convert aerial images to street view images, or vice versa. Compared with these works, our purpose is quite different and requires not only the implicit view projection from front-view to top-view, but also the estimation of road layout and vehicle occupancies under a unified framework.

\textbf{Transformer for vision tasks.}
\wx{Convolutional neural networks(CNNs) are regarded as the most basic component in vision tasks. However, with the recent success of the Transformer~\cite{vaswani2017attention}, its ability to explicitly model pairwise interactions for elements in a sequence has been leveraged in many vision tasks, such as image classification \cite{zhang2020feature}, object detection \cite{carion2020end,zhu2020deformable}, activity recognition \cite{gavrilyuk2020actor}, and image super-resolution \cite{yang2020learning}. ViT \cite{dosovitskiy2021an} first applies the Transformer framework with non-overlapping image patches in the vision task. TNT \cite{han2021transformer} jointly leverages the inner transformer block and outer transformer block to enhance information exchange. Swin Transformer \cite{liu2021swin} obtains a larger receptive field by shifting the windows over the image. PVT \cite{wang2021pyramid} proposes a spatial reduction attention to reduce computational complexity. These models all show even more impressive modeling capabilities and achieve excellent performance. 
Inspired by these transformer-based models, our proposed cross-view transformer} attempts to establish the correlation between the features of views. In addition, we incorporate a feature selection scheme along with the non-local cross-view correlation scheme, significantly enhancing the representativeness of the features.

\section{Our Proposed Method}
\label{sec:Method}

\subsection{Network Overview}

The goal of our work is to estimate the road scene layout and vehicle occupancies on the bird's-eye view in the form of semantic masks given a monocular front-view image. 


\wx{Our network architecture is shown in Fig.~\ref{fig:framework}. The front-view image $I$ is first passed through the encoder based on ResNet~\cite{he2016deep} as the backbone network to extract multi-scale visual features. After that, our proposed front-to-top view projection modules project the encoded features of different scales and propagate them to the decoder to produce the top-view semantic mask $\hat{M}$. In the following subsections, we will elaborate the details of our front-to-top view projection module and its multi-scale deployments on various scales. }

\subsection{Front-to-Top View Projection Module}

Due to the large gap between front-views and top-views, there a lot of image content goes missing during view projection, so the traditional view projection techniques lead to defective results. To this end, the hallucination ability of CNN-based methods has been exploited to address the problem, but the patch-level correlation of both views is not trivial to model within deep networks.

In order to strengthen the view correlation while exploiting the capability of deep networks, we introduce a \wx{front-to-top view projection module into our framework}, which enhances the extracted visual features for projecting front-view to top-view. 
The structure of our proposed \wx{FTVP} module is shown in Fig. \ref{fig:framework}, and it is composed of two parts: cycled view projection and cross-view transformer. 

\textbf{Cycled View Projection (CVP).} Since the features of front-views are not spatially aligned with those of top views due to their large gap, we follow \cite{pan2020cross} and deploy the MLP structure consisting of two fully-connected layers to project the features of front-view to top-view, which can overtake the standard information flow of stacking convolution layers.
As shown in Fig. \ref{fig:framework}, $X$ and $X'$ represent the feature maps before and after view projection, respectively. 
Hence, the holistic view projection can be achieved by: $X' = \mathcal{F}_{MLP}(X)$, where $X$ refers to the features extracted from the ResNet backbone. 
\label{sec:First}

However, such a simple view projection structure cannot guarantee that the information of front-views is effectively delivered. Here, we introduce a cycled self-supervision scheme to consolidate the view projection, which projects the top-view features back to the domain of front-views. 
As illustrated in Fig. \ref{fig:framework}, $X''$ is computed by cycling $X'$ back to the front-view domain via the same MLP structure, i.e., $X'' = \mathcal{F}'_{MLP}(X')$. To guarantee the domain consistency between $X'$ and $X''$, 
we incorporate a cycle loss, i.e., $\mathcal{L}_{cycle}$, as expressed below. 
\begin{align}
    \mathcal{L}_{cycle} &= \| X - X'' \|_1.
\end{align}

\begin{figure}
    \centering
    \begin{tabular}{c@{}c@{}c@{}c@{}c}
    \includegraphics[width=0.13\textwidth,height=2.25cm]{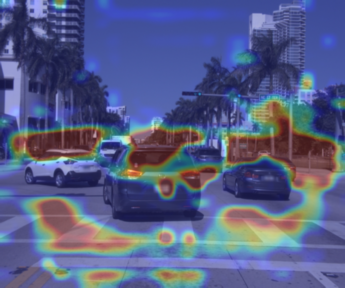}&
    \includegraphics[width=0.005\textwidth,height=2.25cm]{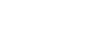}&
    \includegraphics[width=0.13\textwidth,height=2.25cm]{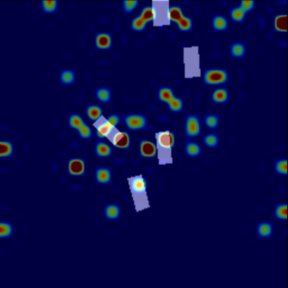}&
    \includegraphics[width=0.005\textwidth,height=2.25cm]{imgs/white.png}&
    \includegraphics[width=0.13\textwidth,height=2.25cm]{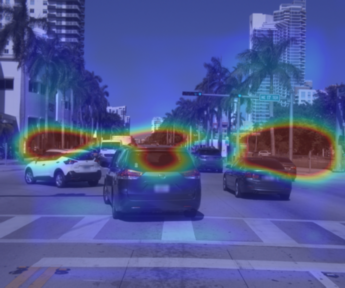}\\
    \includegraphics[width=0.13\textwidth,height=2.25cm]{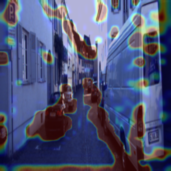}&
    \includegraphics[width=0.005\textwidth,height=2.25cm]{imgs/white.png}&
    \includegraphics[width=0.13\textwidth,height=2.25cm]{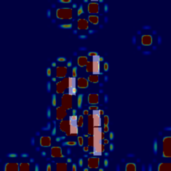}&
    \includegraphics[width=0.005\textwidth,height=2.25cm]{imgs/white.png}&
    \includegraphics[width=0.13\textwidth,height=2.25cm]{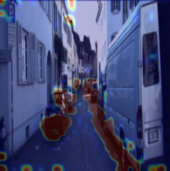}\\
    $X$ & $ $ & $X'$ & $ $ & $X''$
    \end{tabular}
    \caption{Visualization of the features at front-view and top-view by aligning them with the images of the corresponding views.  }
    \label{fig:views}
\end{figure}

The benefits of the cycle structure are two-fold. First, similar to the cycle consistency-based approaches \cite{zhu2017unpaired,dwibedi2019temporal}, cycle loss can innately improve the representativeness of features, since cycling back the top-view features to the front-view domain will strengthen the connection between both views. Second, when the discrepancy between $X$ and $X''$ cannot be further narrowed, $X''$ actually retains the most relevant information for view projection, since $X''$ is reciprocally projected from $X'$. Hence, $X$ and $X'$ refer to the features before and after view projection. $X''$ contains the most relevant features of the front-view for view projection. In Fig.~\ref{fig:views}, we show two examples by visualizing the features of the front-view and top-view. {Specifically, we visualize them by selecting the typical channels of the feature maps (i.e., the 7th and 92nd for two examples of Fig.~\ref{fig:views}) and aligning them with the input images.} As observed, $X$ and $X''$ are similar, but quite different from $X'$ due to the domain difference. We can also observe that, via cycling, $X''$ concentrates more on the road and the vehicles. $X$, $X'$ and $X''$ will be fed into the cross-view transformer. 

\begin{figure}
    \centering
    \includegraphics[width=0.45\textwidth]{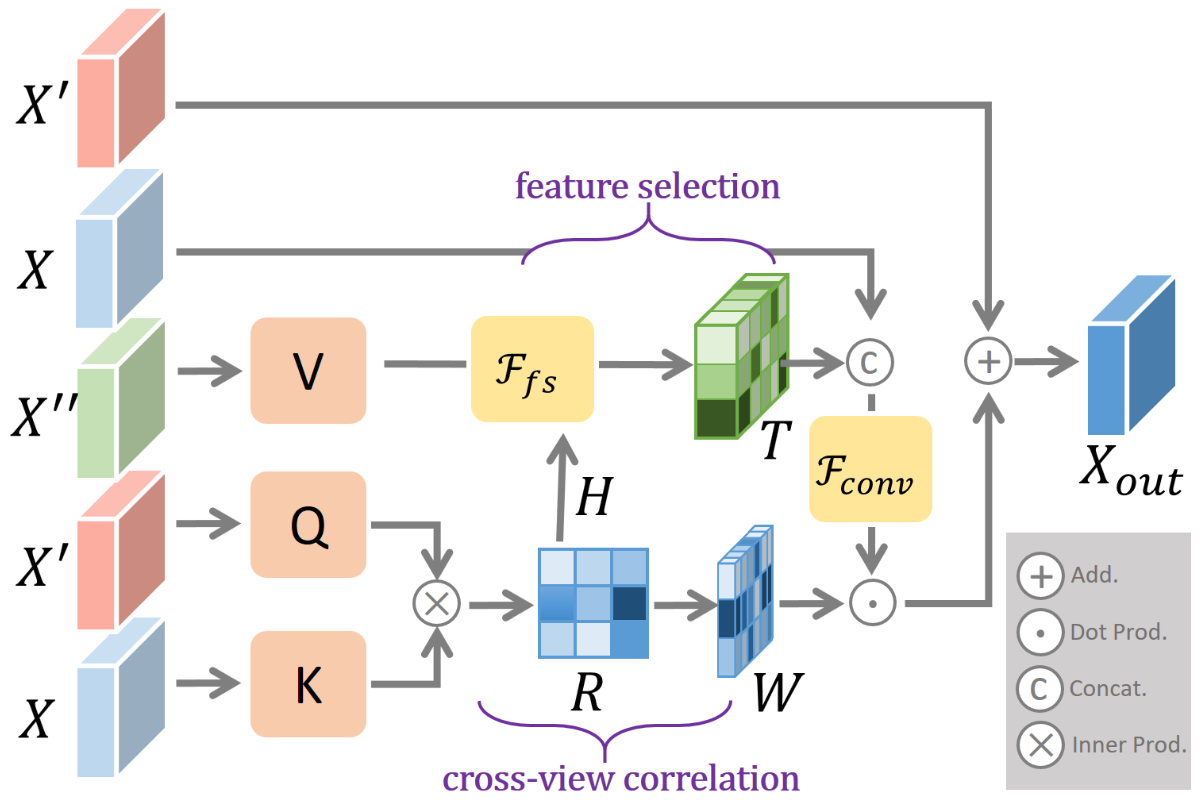}
    \caption{Illustration of cross-view transformer. It contains the cross-view correlation scheme that correlates $X$ and $X'$ to gain the attention map $W$, and the feature selection scheme that extracts the most relevant information from $X''$ to be $T$ \wx{and then uses $X$ to supplement the possibly missing information of $T$}.}
    \label{fig:cvt}
\end{figure}

\textbf{Cross-View Transformer (CVT).} The main purpose of CVT is to correlate the features before view projection (i.e., $X$) and the features after view projection (i.e., $X'$) to strengthen the latter ones. Since $X''$ contains the substantial information of the front-view for view projection, it can be used to further enhance the features as well.
As illustrated in Fig.~\ref{fig:cvt}, CVT can be roughly divided into two schemes: the cross-view correlation scheme that explicitly correlates the features of views to achieve an attention map $W$ to strengthen $X'$ and the feature selection scheme that extracts the most relevant information from $X''$. 

Specifically, $X$, $X'$, and $X''$ serve as the key $K$ ($K \equiv X$), the query $Q$ ($Q \equiv X'$), and the value $V$ ($V \equiv X''$) of CVT, respectively. In our model, the dimensions of $X$, $X'$, and $X''$ are set as the same. 
$X'$ and $X$ are both flattened into patches, and each patch is denoted as $\textbf{x}'_i \in X' (i \in [1,\dots, hw])$ and $\textbf{x}_j \in X (j \in [1,\dots, hw])$, where $hw$ refers to the width of $X$ times its height. 
Thus, the relevance matrix $R$ between any pairwise patches of $X$ and $X'$ can be estimated, i.e., for each patch $\textbf{x}'_i$ in $X'$ and $\textbf{x}_j$ in $X$, their relevance $r_{ij} (\forall r_{ij} \in R)$ is measured by the normalized inner-product:
\begin{align}
    r_{ij} = \langle \frac{\textbf{x}'_i}{||\textbf{x}'_i||}, \frac{\textbf{x}_j}{||\textbf{x}_j||} \rangle.
\end{align}

With the relevance matrix $R$, we create two vectors $W$ ($W=\{w_i\}, \forall i \in [1,\dots, hw]$) and $H$ ($H=\{h_i\},\forall i \in [1,\dots, hw]$) based on the maximum value and the corresponding index for each row of $R$, respectively:
\begin{align}
    w_i &= \max_j r_{ij}, \forall r_{ij} \in R,\\
    h_i &= \arg\max_j r_{ij}, \forall r_{ij} \in R.
\end{align}

Each element of $W$ implies the degree of correlation between each patch of $X'$ and all the patches of $X$, which can serve as an attention map. Each element of $H$ indicates the index of the most relevant patch in $X$ with respect to each patch of $X'$. 

Recall that both $X$ and $X''$ are the features of the front-view, except $X$ contains its complete information, while $X''$ retains the relevant information for view projection. Assuming that the correlation between $X$ and $X'$ is similar to the correlation between $X''$ and $X'$, it is reasonable to utilize the relevance of $X$ and $X'$ (i.e., $R$) to extract the most important information from $X''$. 
To this end, we introduce a feature selection scheme $\mathcal{F}_{fs}$. With $H$ and $X''$, $\mathcal{F}_{fs}$ can produce new feature maps $T$ ($T=\{\textbf{t}_i\},\forall i \in [1,\dots, hw]$) by retrieving the most relevant features from $X''$:
\begin{align}
    \textbf{t}_i = \mathcal{F}_{fs}(X'', h_i), \forall h_i \in H, 
\end{align}
where $\mathcal{F}_{fs}$ retrieves the feature vector $\textbf{t}_i$ from the $h_i$-th position of $X''$.

Hence, $T$ stores the most relevant information of $X''$ for each patch of $X'$. \wx{However, it is worth noting that the information of $T$ all comes from $X''$, which is likely to cause information loss during view projection. To compensate for the lost information, it can be reshaped as the same dimension as $X$ and concatenated with $X$, which supplements the raw information.} Then the concatenated features will be weighted by the attention map $W$ and finally aggregated with $X'$ via a residual structure. To sum up, the process can be formally expressed as:
\begin{align}
    X_{out} = X' + \mathcal{F}_{conv}(\text{Concat}(X, T)) \odot W,
\end{align}
where $\odot$ denotes the element-wise multiplication and $\mathcal{F}_{conv}$ refers to a convolutional layer with $3 \times 3$ kernel size.
$X_{out}$ is the final output of CVT and will then be passed to the decoder network to produce the segmentation mask of the top-view.

\subsection{Multi-scale FTVP Modules}
\label{sec:Second}
\wx{As mentioned above, our proposed FTVP module can acquire the enhanced top-view features $X_{out}$ to generate refined segmentation results. However, the resolution of $X_{out}$ is too low to completely retain the information on visual details, which tends to cause spatial deviation of object location from top-views. }

\wx{To remedy the shortcomings, we introduce multi-scale FTVP modules to bridge the low-level features and the upsampled top-view features in the decoder. As shown in Fig. \ref{fig:framework}, the extracted front-view features $X_i$ of the $i$-th scales are passed through the FTVP modules for view projection in order to gain the corresponding top-view features $\hat{X}_i$. In practice, we employ the three inner most encoded features for projection to avoid unnecessary computational burden. 
Next, the projected features $\hat{X}_i$ and the top-view features $X'_i$ in the upsampling stream are concatenated, i.e., $\hat{X}'_i=\text{Concat}(\hat{X}_i, X'_i)$, so that the projected features $\hat{X}_i$ manage to deliver rich low-level information to the decoder. Finally, we employ the deep supervision scheme to supervise the segmentation results generated by the top-view features $\hat{X}’_i$.}

\begin{figure}
    \centering
    \footnotesize
    
    \begin{tabular}{c}
    \includegraphics[width=0.47\textwidth,height=2.5cm]
        {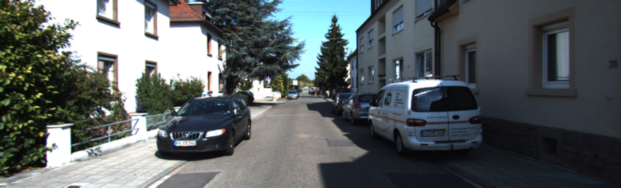} \\
        Front-view Image \\
    \end{tabular}
    
	\begin{tabular}{c@{}c@{}c}
        \includegraphics[width=0.19\textwidth,height=2.5cm]
        {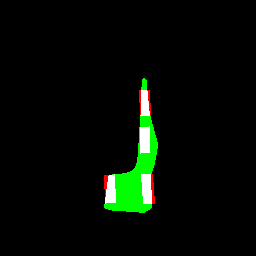} &
		\includegraphics[width=0.14\textwidth,height=2.5cm]
        {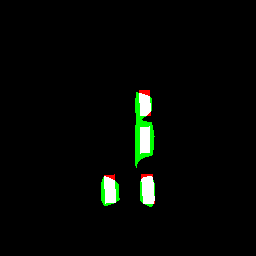} &
		\includegraphics[width=0.14\textwidth,height=2.5cm]
        {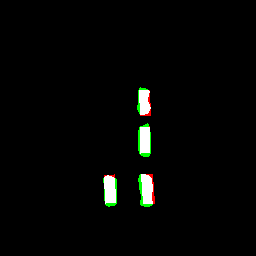} \\
        $i=0$ & $i=1$ & $i=2$ \\
        \includegraphics[width=0.19\textwidth,height=2.5cm]
        {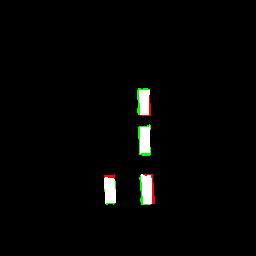} &
		\includegraphics[width=0.14\textwidth,height=2.5cm]
        {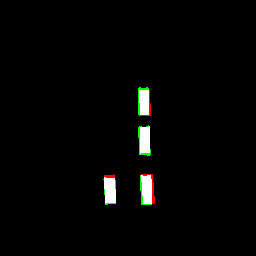} &
		\includegraphics[width=0.14\textwidth,height=2.5cm]
        {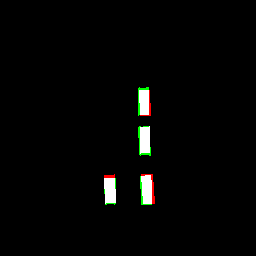} \\
        $i=3$ & $i=4$ & $i=5$
        
	\end{tabular}
	\caption{We visualize the segmentation maps generated from top-view features at the $i$-th scale (i.e., $i \in \{0,...,5\}$). In addition, we highlight the pixels that are discrepancies between our estimation and the ground truth. The green and red pixels represent False Negatives and False Positives, respectively.}
	\label{fig:re}
\end{figure}

\wx{We visualize the segmentation masks generated from the top-view features of different scales in Fig.~\ref{fig:re}. When the resolutions of the features increase, the false negatives (i.e., green areas) and false positives (i.e., red areas) of estimating the locations of vehicles gradually decrease due to the involvement of low-level information. In addition, for the features of $3$-rd, $4$-th, and $5$-th scales (i.e., the three highest resolution intermediate feature maps in the encoder), their produced results are almost the same, so it is sufficient to employ three FTVP modules only.}

\subsection{Loss Function}
Overall, the loss function of our framework is defined as:
\begin{align}
    \mathcal{L} = \sum_{i=0}^{n}\mathcal{L}^{i}_{seg} + \lambda\mathcal{L}_{cycle},
\end{align}
\wx{where $\mathcal{L}^{i}_{seg}$ is the cross-entropy loss for supervising the segmentation results produced by the features of the $i$-th scale. 
The main objective of the network to narrow the gap between the predicted semantic mask and the ground-truth mask. However, real-world scenes often include small objects such as vehicles and pedestrians, which lead to class-imbalance problems. Thus, in practice, we use the square root of the inverse class frequency to weight unbalanced loss for small classes. In addition, $\lambda$ is the balance weights of the cycle loss and it is set as $0.001$.}

\section{Experimental Results}
\label{sec:Results}
To evaluate our proposed model, we conduct several experiments
over a variety of challenging scenarios and compare our results against state-of-the-art methods on public benchmarks. We also perform extensive ablation experiments to delve into our network structure. 

\subsection{Implementation Details}

We implement our framework using Pytorch on a workstation with a single NVIDIA Titan XP GPU card. We adopt ResNet-18~\cite{he2016deep} without bottleneck layers as our backbone. The MLP contains two fully-connected layers and ReLU activation in CVP, and each input of CVT utilizes one convolutional layer with kernel size $1\times 1$. The decoder is composed of 5 convolutional blocks and upsampling layers with the spatial resolution increasing by a factor of 2.
All the input images are normalized to $1024\times 1024$ and the output size is $256 \times 256$.
The network parameters are randomly initialized, and we adopt the Adam optimizer~\cite{kingmaadam} and use a mini-batch size of 6. The initial learning rate is set to $1\times 10^{-4}$ and is decayed by a poly learning rate policy where the initial
learning rate is multiplied by $(1-\frac{iter}{total\_iter})^{0.9}$ after each iteration. 
In practice, it takes 50 epochs to converge our model, and our model can run in real time ($25$ FPS) on our single-GPU platform. 

\begin{figure*}[t]
    \centering
	\begin{tabular}{c@{}c@{}c@{}c@{}c@{}c}
		\includegraphics[width=0.25\textwidth,height=2.1cm]
        {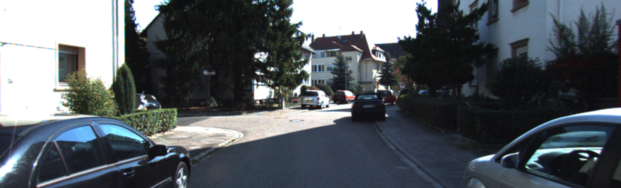} &
		\includegraphics[width=0.15\textwidth,height=2.1cm]
        {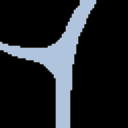}&
		\includegraphics[width=0.15\textwidth,height=2.1cm]
        {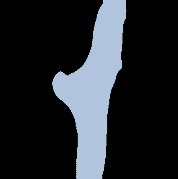}&
        \includegraphics[width=0.15\textwidth,height=2.1cm]
        {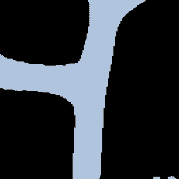}&
        \includegraphics[width=0.15\textwidth,height=2.1cm]
        {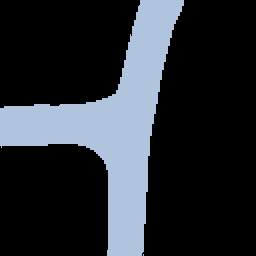}&
		\includegraphics[width=0.15\textwidth,height=2.1cm]
        {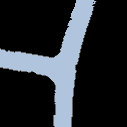}\\
        \includegraphics[width=0.25\textwidth,height=2.1cm]
        {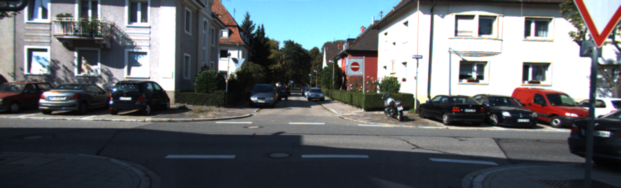} &
		\includegraphics[width=0.15\textwidth,height=2.1cm]
        {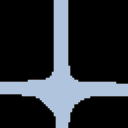}&
		\includegraphics[width=0.15\textwidth,height=2.1cm]
        {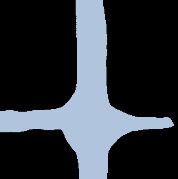}&
        \includegraphics[width=0.15\textwidth,height=2.1cm]
        {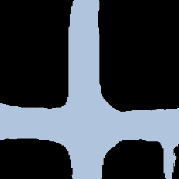}&
        \includegraphics[width=0.15\textwidth,height=2.1cm]
        {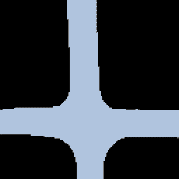}&
		\includegraphics[width=0.15\textwidth,height=2.1cm]
        {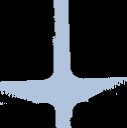}\\
        \includegraphics[width=0.25\textwidth,height=2.1cm]
        {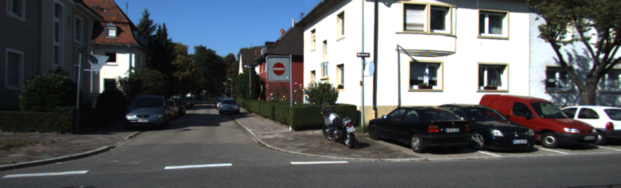} &
		\includegraphics[width=0.15\textwidth,height=2.1cm]
        {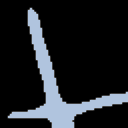}&
		\includegraphics[width=0.15\textwidth,height=2.1cm]
        {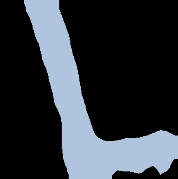}&
        \includegraphics[width=0.15\textwidth,height=2.1cm]
        {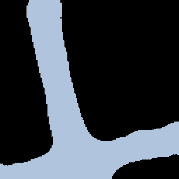}&
        \includegraphics[width=0.15\textwidth,height=2.1cm]
        {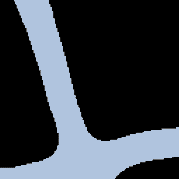}&
		\includegraphics[width=0.15\textwidth,height=2.1cm]
        {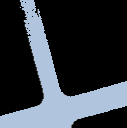}\\
        \includegraphics[width=0.25\textwidth,height=2.1cm]
        {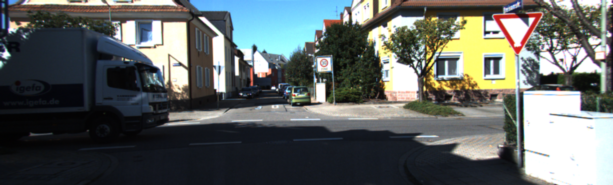} &
		\includegraphics[width=0.15\textwidth,height=2.1cm]
        {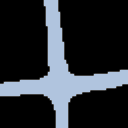}&
		\includegraphics[width=0.15\textwidth,height=2.1cm]
        {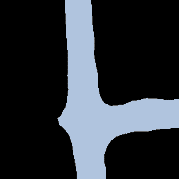}&
        \includegraphics[width=0.15\textwidth,height=2.1cm]
        {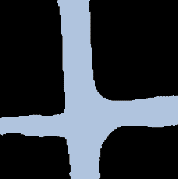}&
        \includegraphics[width=0.15\textwidth,height=2.1cm]
        {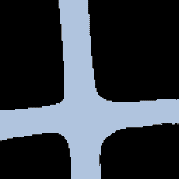}&
		\includegraphics[width=0.15\textwidth,height=2.1cm]
        {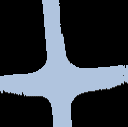}\\
		Front-view & VED & VPN & MonoLayout & Ours & Ground-truth
	\end{tabular}
	\caption{Comparison results of road layout estimation on \textit{KITTI Odometry}.}
	\label{fig:road_layout}
\end{figure*}

\subsection{Datasets}

We evaluate our approach on three datasets: KITTI \cite{geiger2012we}, Argoverse \cite{chang2019argoverse}, and Nuscenes \cite{caesar2020nuscenes}. For performance assessment, we adopt the mean of Intersection-over-Union (mIOU) and Average Precision (mAP) as the evaluation metrics. 

\textbf{KITTI.} Since KITTI has not provided sufficient annotation for road layout or vehicles to be used in our task, we generally follow the practice of \cite{2020MonoLayout}, in which the results are categorized in the following datasets.  
For comparison with state-of-the-art 3D vehicle detection approaches, we evaluate performance on the KITTI 3D object detection (\textit{KITTI 3D Object}) split of Chen et al. \cite{2016Monocular}, i.e., 3712 training images and 3769 validation images. The \textit{KITTI Odometry} dataset is used to evaluate the road layout, the annotation of which comes from the Semantic KITTI dataset \cite{behley2019semantickitti}.
In addition to the previous two datasets, we evaluate the performance on the \textit{KITTI Raw} split used in \cite{2018Learning}, i.e., 10156 training images and 5074 validation images. Since its ground-truths are produced by registering the depth and semantic segmentation of Lidar scans that are not sufficiently dense, we apply image dilation and erosion to produce better ground-truth annotations.

\textbf{Argoverse.} Furthermore, we also compare methods on \textit{Argoverse}, which provides a high-resolution semantic occupancy grid and vehicle detection in top-view for evaluating the spatial layout of roads and vehicles, with 6723 training images and 2418 validation images. \wx{For multiple semantic categories, we follow the evaluation protocol of \cite{roddick2020predicting}, which selects 1 static class (i.e., Drivable) and 7 dynamic object classes (i.e., Vehicle, Pedestrian, Large vehicle, Bicycle, Bus, Trailer, and Motorcycle) to conduct comparison experiments.}

\wx{\textbf{Nuscenes.} The \textit{NuScenes} dataset consists of 1000 twenty-second video clips collected in Boston and Singapore. It provides 3D bounding box annotations for 23 object classes including walkway, car, trailer, and so on. In our experiments, we also follow the protocol of \cite{roddick2020predicting}. We choose 4 static classes (i.e., Drivable, Pedestrian crossing, Walkway, and Carpark) and 10 dynamic object classes (i.e., Car, Truck, Bus, Trailer, Construction vehicle, Pedestrian, Motorcycle, Bicycle, Traffic cone, and Barrier) for evaluation. In addition, we also use their training and validation split, i.e., 28,008 training images and 5,981 validation images.} 

\subsection{Comparison Methods and Performance Evaluation}

\begin{table}[]
\centering
\setlength{\tabcolsep}{4pt}
\caption{Quantitative results on \textit{KITTI Raw} and \textit{KITTI Odometry}.}
\begin{tabular}{@{}c|cc|cc}
\toprule
\multirow{2}{*}{{\color[HTML]{000000} Methods}} & \multicolumn{2}{c}{{\color[HTML]{000000} \textbf{KITTI Raw}}} & \multicolumn{2}{|c}{{\color[HTML]{000000} \textbf{KITTI Odometry}}}            \\ \cmidrule{2-5}
               & {\color[HTML]{000000} mIOU (\%)}   & {\color[HTML]{000000} mAP (\%)}    & {\color[HTML]{000000} mIOU (\%)}           & {\color[HTML]{000000} mAP (\%)}            \\ \midrule
{\color[HTML]{000000} VED \cite{lu2019monocular}}   & {\color[HTML]{000000} 58.41}  & {\color[HTML]{000000} 66.01}      & {\color[HTML]{000000} 65.74}          & {\color[HTML]{000000} 67.84}              \\
{\color[HTML]{000000} VPN \cite{pan2020cross}}   & {\color[HTML]{000000} 59.58}      & {\color[HTML]{000000} 79.07}      & {\color[HTML]{000000} 66.81}              & {\color[HTML]{000000} 81.79}              \\
{\color[HTML]{000000} MonoLayout~\cite{2020MonoLayout}}      & {\color[HTML]{000000} 66.02}      & {\color[HTML]{000000} 75.73}      & {\color[HTML]{000000} 76.15}          & {\color[HTML]{000000} 85.25}          \\
{\color[HTML]{000000} CVT \cite{yang2021projecting}}            & {\color[HTML]{000000} 68.34}  & {\color[HTML]{000000} 80.78}  & {\color[HTML]{000000} 77.49} & {\color[HTML]{000000} 86.69} \\
Ours & \textbf{68.42} & \textbf{81.78} & \textbf{77.66} & \textbf{87.82} \\
\bottomrule
\end{tabular}
\label{tab:kitti_sota}
\end{table}

\begin{table}[]
\centering
\setlength{\tabcolsep}{4pt}
\caption{Comparison results on \textit{Argoverse Road} and \textit{Argoverse Vehicle}.}
\begin{tabular}{@{}c|cc|cc}
\toprule
\multirow{2}{*}{{\color[HTML]{000000} Methods}}            & \multicolumn{2}{c}{{\color[HTML]{000000} \textbf{Road}}}                 & \multicolumn{2}{|c}{{\color[HTML]{000000} \textbf{Vehicle}}}              \\ \cmidrule{2-5}
              & {\color[HTML]{000000} mIOU (\%)}           & {\color[HTML]{000000} mAP (\%)}            & {\color[HTML]{000000} mIOU (\%)}           & {\color[HTML]{000000} mAP (\%)}            \\ \midrule
{\color[HTML]{000000} VED \cite{lu2019monocular}} & {\color[HTML]{000000} {72.84}}          & {\color[HTML]{000000} {78.11}}              & {\color[HTML]{000000} 24.16}          & {\color[HTML]{000000} 36.83}          \\
{\color[HTML]{000000} VPN \cite{pan2020cross}} & {\color[HTML]{000000} 71.07}          & {\color[HTML]{000000} 86.83}          & {\color[HTML]{000000} 16.58}          & {\color[HTML]{000000} 39.73}          \\
{\color[HTML]{000000} MonoLayout~\cite{2020MonoLayout}}    & {\color[HTML]{000000} 73.25}          & {\color[HTML]{000000} 84.56}          & {\color[HTML]{000000} 32.58}          & {\color[HTML]{000000} 51.06}          \\
{\color[HTML]{000000} CVT \cite{yang2021projecting}}          & {\color[HTML]{000000} 76.51} & {\color[HTML]{000000} 87.21} & {\color[HTML]{000000} \textbf{48.48}} & {\color[HTML]{000000} 64.04} \\
Ours & \textbf{78.14} & \textbf{88.05} & 47.71 & \textbf{66.33} \\
\bottomrule
\end{tabular}
\label{tab:argo_sota}
\end{table}
\begin{table}[]
\centering
\setlength{\tabcolsep}{12pt}
\caption{Results on \textit{KITTI 3D Object}.}
\begin{tabular}{c|cc}
\toprule
{\color[HTML]{000000} Methods}              & {\color[HTML]{000000} {mIOU (\%)}}  & {\color[HTML]{000000} {mAP (\%)}}   \\ \midrule
{\color[HTML]{000000} VED \cite{lu2019monocular}} & {\color[HTML]{000000} 20.45}          & {\color[HTML]{000000} 22.59}          \\
{\color[HTML]{000000} Mono3D~\cite{2016Monocular}}        & {\color[HTML]{000000} 17.11}          & {\color[HTML]{000000} 26.62}          \\
{\color[HTML]{000000} OFT~\cite{roddick2018orthographic}}           & {\color[HTML]{000000} 25.24}          & {\color[HTML]{000000} 34.69}          \\
{\color[HTML]{000000} VPN \cite{pan2020cross}} & {\color[HTML]{000000} 16.80}              & {\color[HTML]{000000} 35.54}              \\
{\color[HTML]{000000} MonoLayout~\cite{2020MonoLayout}}    & {\color[HTML]{000000} 30.18}          & {\color[HTML]{000000} 45.91}          \\
{\color[HTML]{000000} CVT\cite{yang2021projecting}}          & {\color[HTML]{000000} 38.79} & {\color[HTML]{000000} 50.26} \\
Ours & \textbf{40.69} & \textbf{59.05} \\
\bottomrule
\end{tabular}
\label{tab:object_sota}
\end{table}

\textbf{Comparison Methods.} 
For evaluation, we compare our model with some of the state-of-the-art methods \wx{for road layout estimation and vehicle occupancy estimation}, including VED~\cite{lu2019monocular}, MonoLayout~\cite{2020MonoLayout}, VPN~\cite{pan2020cross}, Mono3D~\cite{2016Monocular}, OFT~\cite{roddick2018orthographic}, and CVT \cite{yang2021projecting}. Among these methods, Mono3D~\cite{2016Monocular} and OFT~\cite{roddick2018orthographic} are specifically used to detect vehicles in top-view. 
For the quantitative results, we follow the ones reported in \cite{2020MonoLayout}. For MonoLayout~\cite{2020MonoLayout}, we compare with their latest online reported results, which are generally better than the ones reported in their original paper.
VPN~\cite{pan2020cross} originally adopted multiple views from different cameras to generate the top-view representation. We adapt their model for single-view input and then retrain it using the same training protocol for our task. Likewise, VED~\cite{lu2019monocular} is retrained for the benchmarks of road layout estimation, and we obtain comparable or better results than the ones reported in \cite{2020MonoLayout}. \wx{CVT \cite{yang2021projecting} is our preliminary version, which correlates the features of the views before and after projection and performs feature selection.}

\wx{In addition, we also compare our model against the state-of-the-art methods for multi-class semantic estimation. These methods include VED \cite{lu2019monocular}, PointPillars \cite{lang2019pointpillars}, VPN \cite{pan2020cross}, PON \cite{roddick2020predicting}, OFT \cite{roddick2018orthographic}, 2D-Lift \cite{dwivedi2021bird}, Stitch \cite{can2022understanding}, and STA \cite{saha2021enabling}. Most of these works focus on estimating the top-view semantic segmentation maps, and the rest use detection methods. We follow the single frame quantitative results reported in \cite{dwivedi2021bird}, \cite{can2022understanding}, and \cite{saha2021enabling} for the \textit{Argoverse} and \textit{Nuscenes} datasets.
}

\textbf{Road layout estimation.} To evaluate the performance of our model on the task of road layout estimation, we compare our model against VED~\cite{lu2019monocular}, MonoLayout~\cite{2020MonoLayout}, VPN~\cite{pan2020cross}, and CVT~\cite{yang2021projecting} on the \textit{KITTI Raw} and \textit{KITTI Odometry} datasets. Note that, since we post-process the ground-truth annotations of \textit{KITTI Raw}, we retrain all the comparison methods under the same training protocol.
The comparison results are demonstrated in Table \ref{tab:kitti_sota}. Additionally, we also compare them on \textit{Argoverse Road}, as shown in Table \ref{tab:argo_sota}. 
As observed, in these three benchmarks, our model shows advantages over the competitors in both mIOU and mAP. \wx{In addition, compared with our preliminary version (i.e., CVT~\cite{yang2021projecting}), the performance also exceeds about +1\%, especially in mAP, because the spatial deviation has been rectified using multi-scale FTVP.} Examples are shown in Fig.~\ref{fig:road_layout}. Note that the ground-truths may contain noise, since they are converted from the Lidar measurements. Even so, our approach can still produce satisfactory results. 

\begin{figure}
    
    \footnotesize
    \vspace{-0.3cm}
	\begin{tabular}{c@{}c@{}c@{}c@{}c}
        \includegraphics[width=0.14\textwidth,height=1.9cm]
        {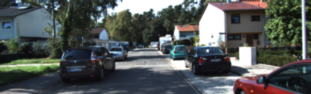} &
		\includegraphics[width=0.11\textwidth,height=1.9cm]
        {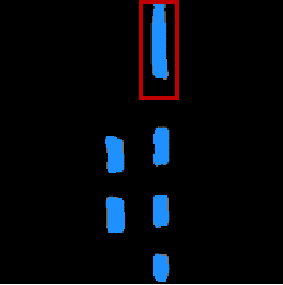}&
		\includegraphics[width=0.11\textwidth,height=1.9cm]
        {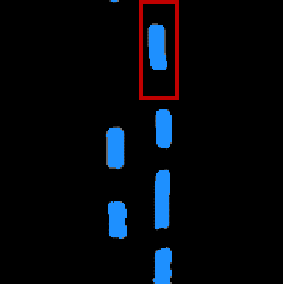}&
		\includegraphics[width=0.11\textwidth,height=1.9cm]
        {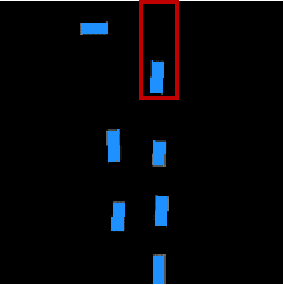}\\
        \includegraphics[width=0.14\textwidth,height=1.9cm]
        {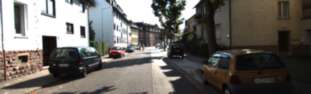} &
		\includegraphics[width=0.11\textwidth,height=1.9cm]
        {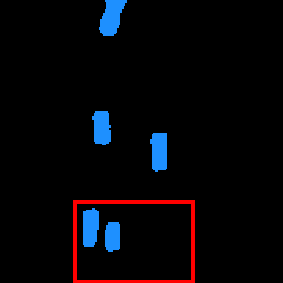}&
		\includegraphics[width=0.11\textwidth,height=1.9cm]
        {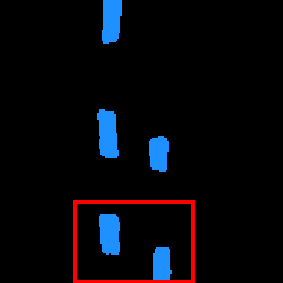}&
		\includegraphics[width=0.11\textwidth,height=1.9cm]
        {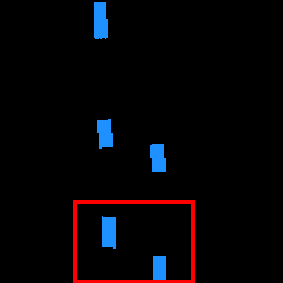}\\
        \includegraphics[width=0.14\textwidth,height=1.9cm]
        {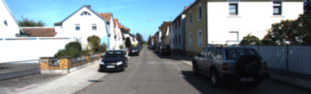} &
		\includegraphics[width=0.11\textwidth,height=1.9cm]
        {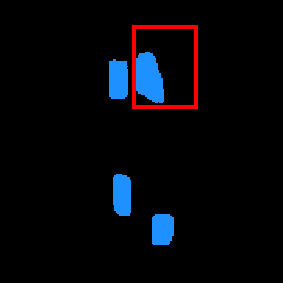}&
		\includegraphics[width=0.11\textwidth,height=1.9cm]
        {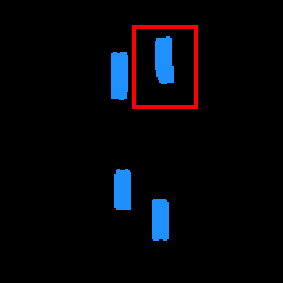}&
		\includegraphics[width=0.11\textwidth,height=1.9cm]
        {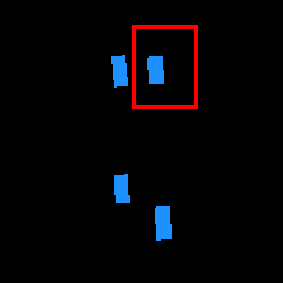}\\
        \includegraphics[width=0.14\textwidth,height=1.9cm]
        {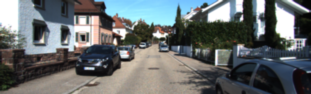} &
		\includegraphics[width=0.11\textwidth,height=1.9cm]
        {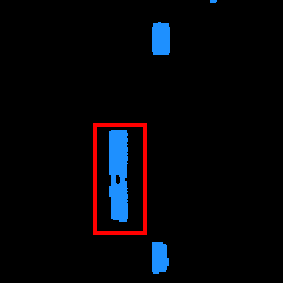}&
		\includegraphics[width=0.11\textwidth,height=1.9cm]
        {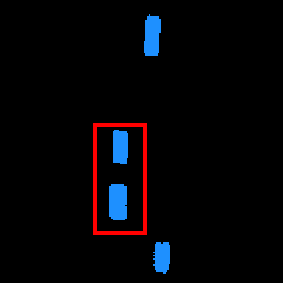}&
		\includegraphics[width=0.11\textwidth,height=1.9cm]
        {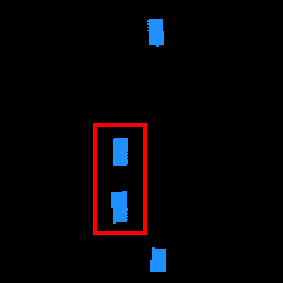}\\
        \includegraphics[width=0.14\textwidth,height=1.9cm]
        {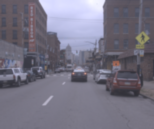} &
		\includegraphics[width=0.11\textwidth,height=1.9cm]
        {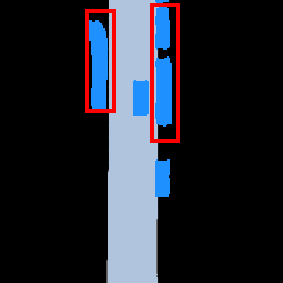}&
		\includegraphics[width=0.11\textwidth,height=1.9cm]
        {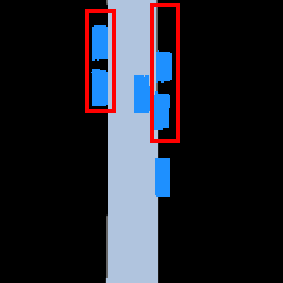}&
		\includegraphics[width=0.11\textwidth,height=1.9cm]
        {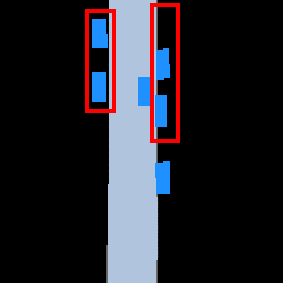}\\
        \includegraphics[width=0.14\textwidth,height=1.9cm]
        {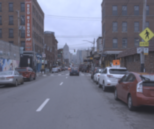} &
		\includegraphics[width=0.11\textwidth,height=1.9cm]
        {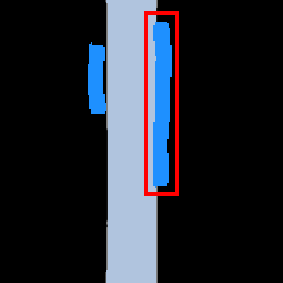}&
		\includegraphics[width=0.11\textwidth,height=1.9cm]
        {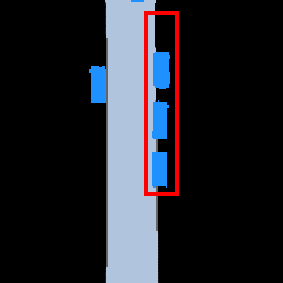}&
		\includegraphics[width=0.11\textwidth,height=1.9cm]
        {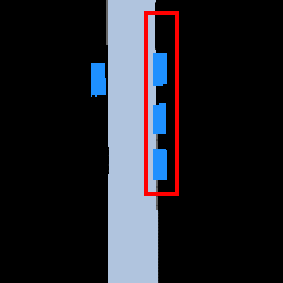}\\
        \includegraphics[width=0.14\textwidth,height=1.9cm]
        {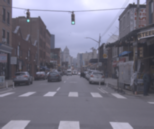} &
		\includegraphics[width=0.11\textwidth,height=1.9cm]
        {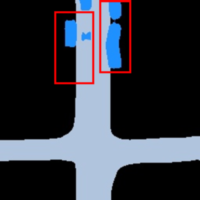}&
		\includegraphics[width=0.11\textwidth,height=1.9cm]
        {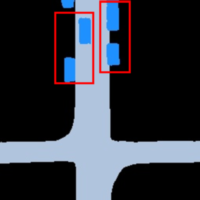}&
		\includegraphics[width=0.11\textwidth,height=1.9cm]
        {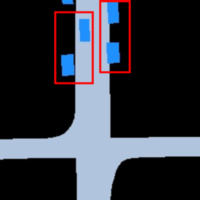}\\
        \includegraphics[width=0.14\textwidth,height=1.9cm]
        {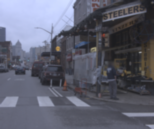} &
		\includegraphics[width=0.11\textwidth,height=1.9cm]
        {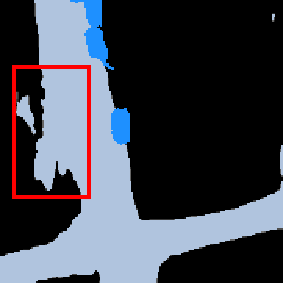}&
		\includegraphics[width=0.11\textwidth,height=1.9cm]
        {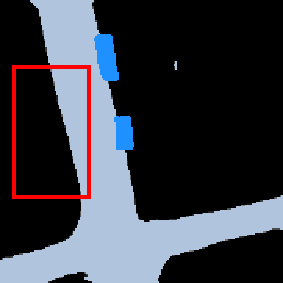}&
		\includegraphics[width=0.11\textwidth,height=1.9cm]
        {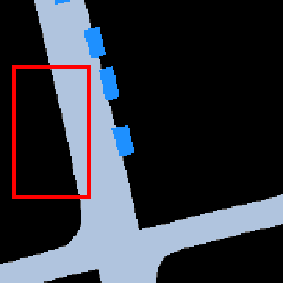}\\
		Front-view  & MonoLayout & Ours & Ground-truth
	\end{tabular}
	\caption{Vehicle occupancy estimation results on \textit{KITTI 3D Object} and the joint estimation on \textit{Argoverse}.}
	\label{fig:vehicle_results}
\end{figure}

\textbf{Vehicle occupancy estimation.} Compared with road layout estimation, estimating vehicle occupancies is a more challenging task, since the scales of vehicles vary and there exist mutual occlusions in the scenes. For evaluation, we perform comparison experiments on the \textit{KITTI 3D Object} and \textit{Argoverse Vehicle} benchmarks against VED \cite{lu2019monocular}, Mono3D~\cite{2016Monocular}, OFT~\cite{roddick2018orthographic}, MonoLayout~\cite{2020MonoLayout}, VPN \cite{pan2020cross}, and CVT~\cite{yang2021projecting}. The results are shown in Tables \ref{tab:argo_sota} and \ref{tab:object_sota}. In Table \ref{tab:object_sota}, our model demonstrates superior performance against the comparison methods. Since \textit{KITTI 3D Object} contains several challenging scenarios, most comparison methods barely obtain $30\%$ mIOU and $50\%$ mAP, while our model gains $40.69\%$ and $59.07\%$, respectively, which shows at least $34.8\%$ and $28.7\%$ improvement over prior methods. \lqq{Our model achieves significant advantage over CVT by at least $4.9\%$ and $17.5\%$ in mIOU and mAP, brought by the structure improvement of FTVP as well as the multi-scale FTVP modules, which beefs up the model and corrects the spatial deviation.}
For the evaluation on \textit{Argoverse Vehicle}, our model outperforms others by a large margin, i.e., at least $46.4\%$ and $29.9\%$ boost over the comparison methods in mIOU and mAP, respectively. 
On the first three rows of Fig.~\ref{fig:vehicle_results}, we show the examples on vehicle occupancy estimation on \textit{KITTI 3D Object}. For the challenging cases with multiple vehicles parked on the sides of roads, our model can still perform well. The last four rows of Fig.~\ref{fig:vehicle_results} show examples of the joint estimation for roads and vehicles on \textit{Argoverse}, and we highlight the advantages of our results.

\begin{table*}[]
    \centering
    \setlength{\tabcolsep}{3.5pt}
    \caption{Comparison results on \textit{Nuscenes}. \wxr{Note that all methods take a single frame as input for fair comparison.}}
    \begin{tabular}{c|c|c|c|c|c|c|c|c|c|c|c|c|c|c|c}
        \toprule
        Model & Driva. & Ped. cross. & Walkway & Carpark & Car & Truck & Bus & Trail. & Constr. veh. & Ped. & Motor. & Bicyc. & Traf. cone & Barrier & Mean \\
        \midrule
        VED \cite{lu2019monocular} & 54.7 & 12.0 & 20.7 & 13.5 & 8.8 & 0.2 & 0.0 & 7.4 & 0.0 & 0.0 & 0.0 & 0.0 & 0.0 & 4.0 & 8.7 \\
        PointPillars \cite{lang2019pointpillars} & 58.1 & 26.0 & 27.4 & 15.7 & 23.9 & 14.6 & 17.6 & 13.2 & 3.0 & 2.3 & 3.8 & 3.4 & 3.6 & 7.1 & 15.8 \\
        VPN \cite{pan2020cross} & 58.0 & 27.3 & 29.4 & 12.9 & 25.5 & 17.3 & 20.0 & 16.6 & 4.9 & 7.1 & 5.6 & 4.4 & 4.6 & 10.8 & 17.5 \\
        PON \cite{roddick2020predicting} & 60.4 & 28.0 & 31.0 & 18.4 & 24.7 & 16.8 & 20.8 & 16.6 & 12.3 & 8.2 & 7.0 & 9.4 & 5.7 & 8.1 & 19.1 \\
        OFT \cite{roddick2018orthographic} & 62.4 & 30.9 & 34.5 & 23.5 & 34.7 & 17.4 & 23.2 & 18.2 & 3.7 & 1.2 & 6.6 & 4.6 & 1.1 & 12.9 & 19.6 \\
        2D-Lift \cite{dwivedi2021bird} & 62.3 & 31.8 & 37.3 & 25.2 & 37.4 & 18.7 & 24.8 & 16.4 & 4.7 & 3.4 & 7.9 & 7.2 & 3.9 & 13.6 & 21.0 \\
        Stitch \cite{can2022understanding} & 71.7 & 27.2 & 34.9 & 32.1 & 32.9 & 15.3 & 23.1 & 15.2 & 4.4 & 5.8 & 8.3 & 6.8 & 4.8 & 17.7 & 21.4 \\
        STA \cite{saha2021enabling} & 71.1 & 31.5 & 32.0 & 28.0 & 34.6 & 18.1 & 22.8 & 11.4 & 10.0 & 7.4 & 7.1 & 14.6 & 5.8 & 10.8 & 21.8 \\
        Ours & 73.4 & 36.1 & 38.7 & 30.8 & 34.6 & 19.0 & 25.5 & 15.9 & 1.6 & 4.8 & 7.2 & 7.1 & 4.4 & 15.9 & 22.5 \\
        \bottomrule
    \end{tabular}
    \label{tab:nu}
\end{table*}

\begin{table}[]
    \centering
    \setlength{\tabcolsep}{2.5pt}
    \caption{Comparison results on \textit{Argoverse}. \wxr{We follow the quantitative results of prior works reported in \cite{can2022understanding}.}}
    \begin{tabular}{c|c|c|c|c|c|c|c|c|c}
        \toprule
        Model & Driva. & Veh. & Ped. & Lar. veh. & Bicyc. & Bus & Trail. & Motor. & Mean \\
        \midrule
        VED \cite{lu2019monocular} & 62.9 & 14.0 & 1.0 & 3.9 & 0.0 & 12.3 & 1.3 & 0.0 & 11.9 \\
        VPN \cite{pan2020cross} & 64.9 & 23.9 & 6.2 & 9.7 & 0.9 & 3.0 & 0.4 & 1.9 & 13.9 \\
        PON \cite{roddick2020predicting} & 65.4 & 31.4 & 7.4 & 11.1 & 3.6 & 11.0 & 0.7 & 5.7 & 17.0 \\
        Stitch \cite{can2022understanding} & 79.8 & 28.2 & 4.8 & 11.4 & 0.0 & 22.5 & 1.1 & 0.4 & 18.5 \\
        Ours & 82.6 & 36.1 & 3.5 & 15.2 & 0.4 & 0.3 & 62.3 & 0.2 & 25.1 \\
        \bottomrule
    \end{tabular}
    \label{tab:agro}
\end{table}

\begin{figure*}
    \centering
	\begin{tabular}{c@{}c@{}c@{}c@{}c@{}c@{}c}
	
        \includegraphics[width=0.14\textwidth,height=2.4cm]
        {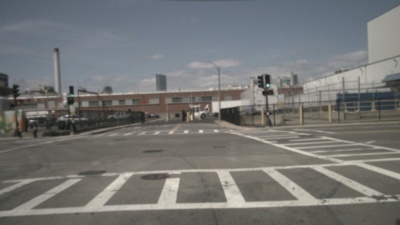} &
        \includegraphics[width=0.14\textwidth,height=2.4cm]
        {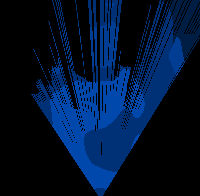} &
		\includegraphics[width=0.14\textwidth,height=2.4cm]
        {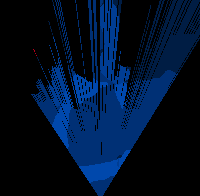}&
		\includegraphics[width=0.14\textwidth,height=2.4cm]
        {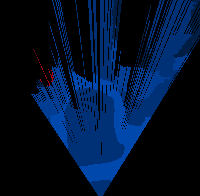}&
        \includegraphics[width=0.14\textwidth,height=2.4cm]
        {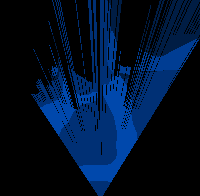}&
        \includegraphics[width=0.14\textwidth,height=2.4cm]
        {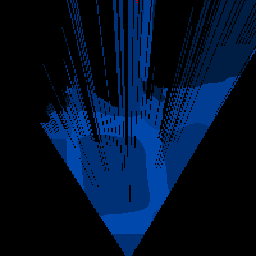}&
		\includegraphics[width=0.14\textwidth,height=2.4cm]
        {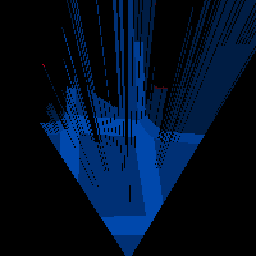}\\
        
        \includegraphics[width=0.14\textwidth,height=2.4cm]
        {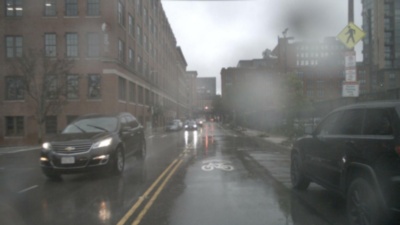} &
        \includegraphics[width=0.14\textwidth,height=2.4cm]
        {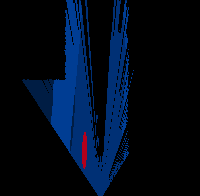} &
		\includegraphics[width=0.14\textwidth,height=2.4cm]
        {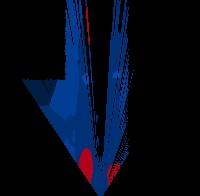}&
		\includegraphics[width=0.14\textwidth,height=2.4cm]
        {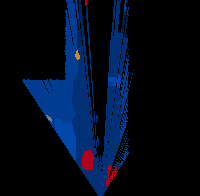}&
        \includegraphics[width=0.14\textwidth,height=2.4cm]
        {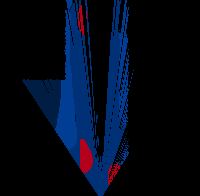}&
        \includegraphics[width=0.14\textwidth,height=2.4cm]
        {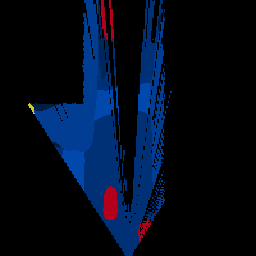}&
		\includegraphics[width=0.14\textwidth,height=2.4cm]
        {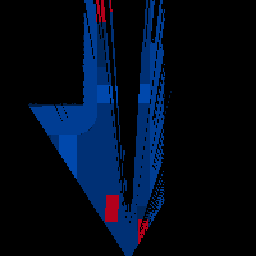}\\
        
        \includegraphics[width=0.14\textwidth,height=2.4cm]
        {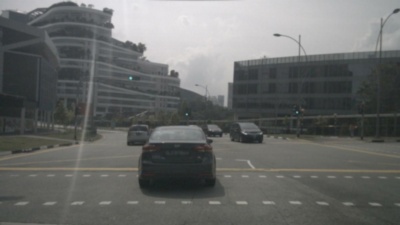} &
        \includegraphics[width=0.14\textwidth,height=2.4cm]
        {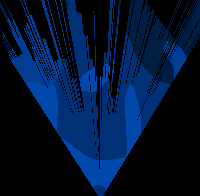} &
		\includegraphics[width=0.14\textwidth,height=2.4cm]
        {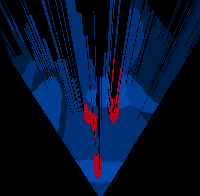}&
		\includegraphics[width=0.14\textwidth,height=2.4cm]
        {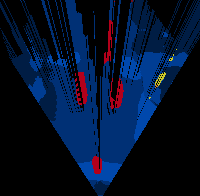}&
        \includegraphics[width=0.14\textwidth,height=2.4cm]
        {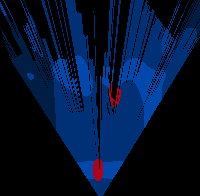}&
        \includegraphics[width=0.14\textwidth,height=2.4cm]
        {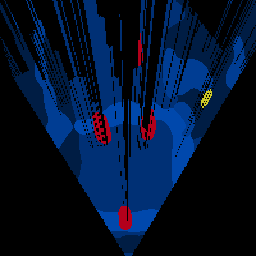}&
		\includegraphics[width=0.14\textwidth,height=2.4cm]
        {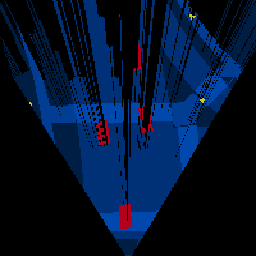}\\
        
        \includegraphics[width=0.14\textwidth,height=2.4cm]
        {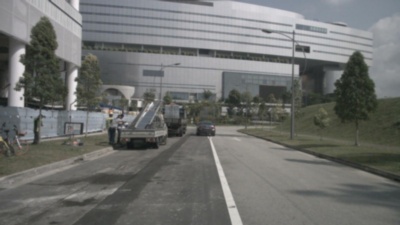} &
        \includegraphics[width=0.14\textwidth,height=2.4cm]
        {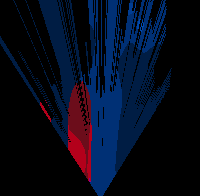} &
		\includegraphics[width=0.14\textwidth,height=2.4cm]
        {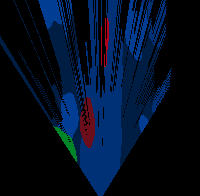}&
		\includegraphics[width=0.14\textwidth,height=2.4cm]
        {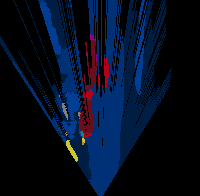}&
        \includegraphics[width=0.14\textwidth,height=2.4cm]
        {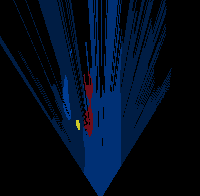}&
        \includegraphics[width=0.14\textwidth,height=2.4cm]
        {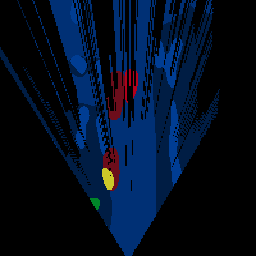}&
		\includegraphics[width=0.14\textwidth,height=2.4cm]
        {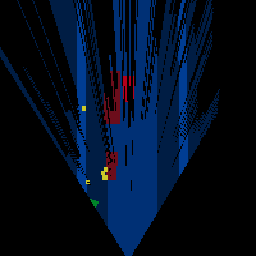}\\

        \includegraphics[width=0.14\textwidth,height=2.4cm]
        {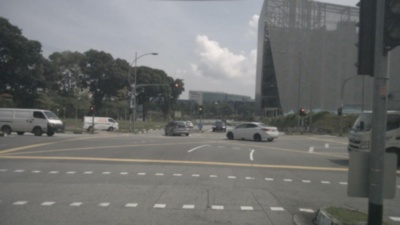} &
        \includegraphics[width=0.14\textwidth,height=2.4cm]
        {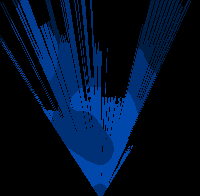} &
		\includegraphics[width=0.14\textwidth,height=2.4cm]
        {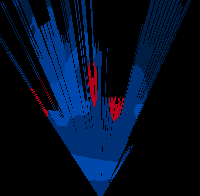}&
		\includegraphics[width=0.14\textwidth,height=2.4cm]
        {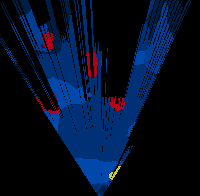}&
        \includegraphics[width=0.14\textwidth,height=2.4cm]
        {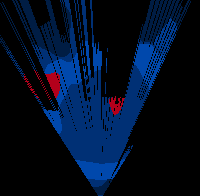}&
        \includegraphics[width=0.14\textwidth,height=2.4cm]
        {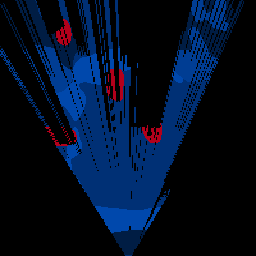}&
		\includegraphics[width=0.14\textwidth,height=2.4cm]
        {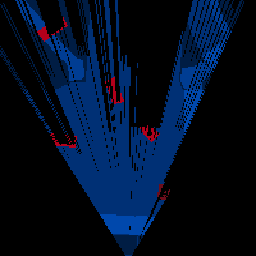}\\
        
        \includegraphics[width=0.14\textwidth,height=2.4cm]
        {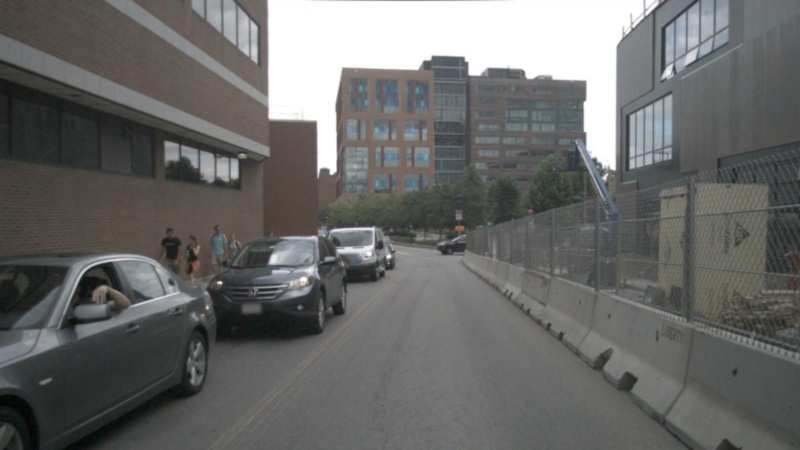} &
        \includegraphics[width=0.14\textwidth,height=2.4cm]
        {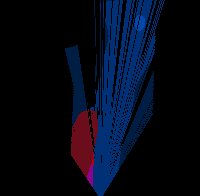} &
		\includegraphics[width=0.14\textwidth,height=2.4cm]
        {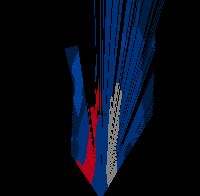}&
		\includegraphics[width=0.14\textwidth,height=2.4cm]
        {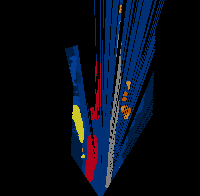}&
        \includegraphics[width=0.14\textwidth,height=2.4cm]
        {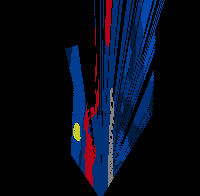}&
        \includegraphics[width=0.14\textwidth,height=2.4cm]
        {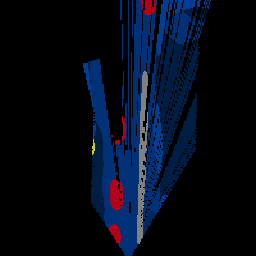}&
		\includegraphics[width=0.14\textwidth,height=2.4cm]
        {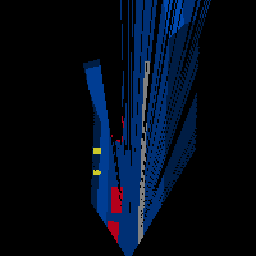}\\
        
        \includegraphics[width=0.14\textwidth,height=2.4cm]
        {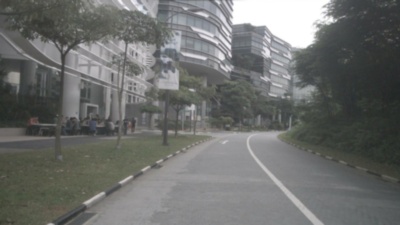} &
        \includegraphics[width=0.14\textwidth,height=2.4cm]
        {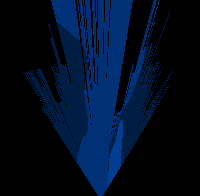} &
		\includegraphics[width=0.14\textwidth,height=2.4cm]
        {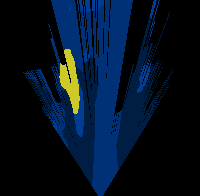}&
		\includegraphics[width=0.14\textwidth,height=2.4cm]
        {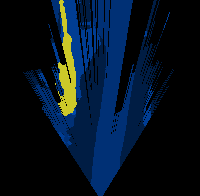}&
        \includegraphics[width=0.14\textwidth,height=2.4cm]
        {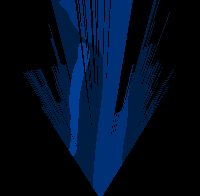}&
        \includegraphics[width=0.14\textwidth,height=2.4cm]
        {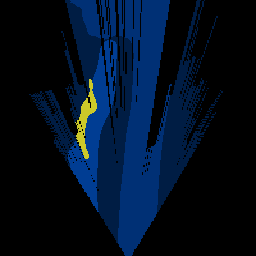}&
		\includegraphics[width=0.14\textwidth,height=2.4cm]
        {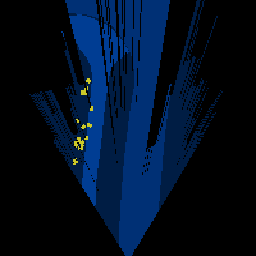}\\
        
        \includegraphics[width=0.14\textwidth,height=2.4cm]
        {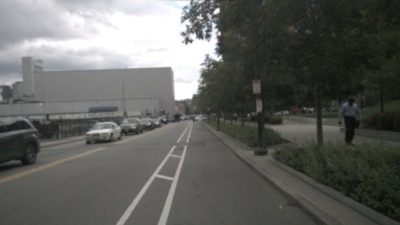} &
        \includegraphics[width=0.14\textwidth,height=2.4cm]
        {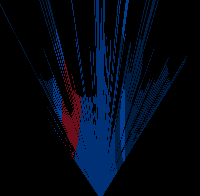} &
		\includegraphics[width=0.14\textwidth,height=2.4cm]
        {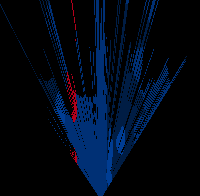}&
		\includegraphics[width=0.14\textwidth,height=2.4cm]
        {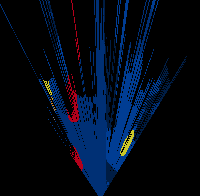}&
        \includegraphics[width=0.14\textwidth,height=2.4cm]
        {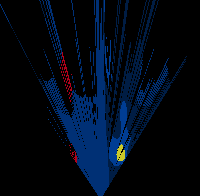}&
        \includegraphics[width=0.14\textwidth,height=2.4cm]
        {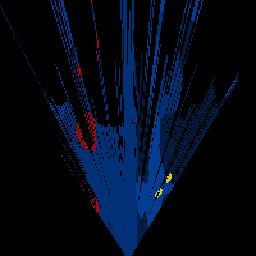}&
		\includegraphics[width=0.14\textwidth,height=2.4cm]
        {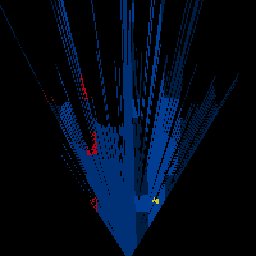}\\
        
        \includegraphics[width=0.14\textwidth,height=2.4cm]
        {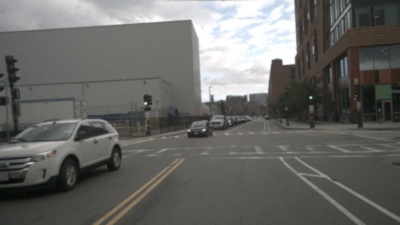} &
        \includegraphics[width=0.14\textwidth,height=2.4cm]
        {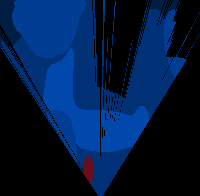} &
		\includegraphics[width=0.14\textwidth,height=2.4cm]
        {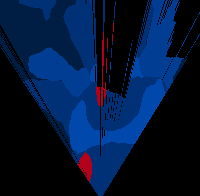}&
		\includegraphics[width=0.14\textwidth,height=2.4cm]
        {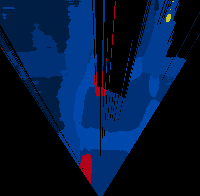}&
        \includegraphics[width=0.14\textwidth,height=2.4cm]
        {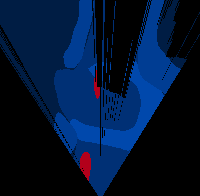}&
        \includegraphics[width=0.14\textwidth,height=2.4cm]
        {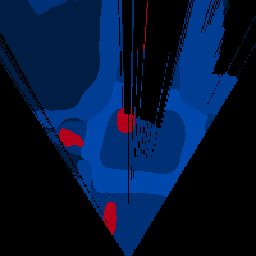}&
		\includegraphics[width=0.14\textwidth,height=2.4cm]
        {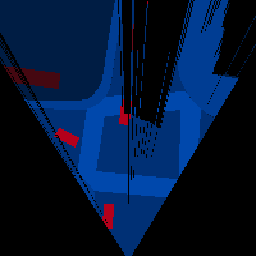}\\
        
		Front-view & VED & VPN & PON & Stitch & Ours & Ground-truth
		
	\end{tabular}
	\caption{\wxr{Examples of static and dynamic object location estimation on \textit{Nuscenes}.}}
	\label{fig:fig_nu}
\end{figure*}

\textbf{Multi-class semantic estimation.} \wx{We extend our model to address multi-class semantic segmentation problems on \textit{Argoverse} with 8 classes and on \textit{Nuscenes} with 14 classes. Compared with road layout estimation and vehicle occupancy estimation, multi-class semantic estimation is a more challenging task because it must accommodate the broad roads and a wide variety of vehicles with different scales. We conduct comparison experiments against VED \cite{lu2019monocular}, VPN \cite{pan2020cross}, PON \cite{roddick2020predicting}, and Stitch \cite{can2022understanding} for the \textit{Argoverse} dataset. For \textit{Nuscenes}, we introduce more comparison methods including PointPillars \cite{lang2019pointpillars}, OFT \cite{roddick2018orthographic}, 2D-Lift \cite{dwivedi2021bird}, and STA \cite{saha2021enabling}. The comparison results are depicted in Tables~\ref{tab:nu} and \ref{tab:agro}. As observed, our model shows superior performance over other methods. Our model offers significant improvement over the state-of-the-art methods on \textit{Argoverse} for the ``trailer" class.}
\wxr{In Fig.~\ref{fig:fig_nu}, we illustrate the examples of multi-class semantic estimation on \textit{Nuscenes}. We observe
that our model performs better than other methods on position and shape estimation for both static and dynamic classes.}

\label{sec:Third}

\subsection{Ablation Study}

\label{sec:Fourth}
To delve into our network structure, we conduct several ablation experiments using the front-to-top view projection module and the context-aware discriminator. 

\textbf{Front-to-top view projection module.} Recall that our front-to-top view projection module consists of CVP and CVT. Specifically, CVP can be divided into the MLP and the cycle structure. CVT can be decomposed into a cross-view correlation scheme and a feature selection scheme. 
In the following, we will investigate the necessity of these modules based on the dataset \textit{KITTI 3D Object} in Table \ref{tab:cvt1}. 

First, the baseline is the vanilla encoder-decoder network using the same encoder and decoder as our model. Then we insert the MLP structure to the baseline. As shown in Table \ref{tab:cvt1}, it obviously improves the effectiveness of view projection. Next, we add a cross-view correlation scheme into the network, which measures the correlation of $X$ and $X'$ and applies it as the attention map to enhance $X'$. As observed, with the involvement of the cross-view correlation scheme, the performance is significantly boosted. We then introduce the cycle structure as well as the cycle loss into the network, in which $X''$ will be fed into the cross-view correlation scheme. Finally, we insert the feature selection scheme, which further strengthens the performance of the model. 

\begin{figure*}
    \centering
    \footnotesize
    \begin{tabular}{c@{}c@{}c@{}c@{}c@{}c@{}c}
        \includegraphics[width=0.22\textwidth,height=1.9cm]
        {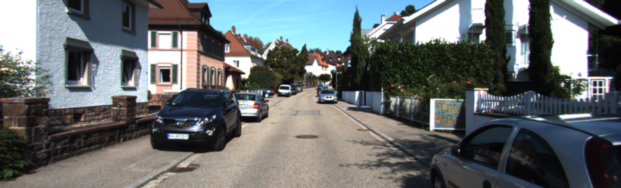} &
		\includegraphics[width=0.13\textwidth,height=1.9cm]
        {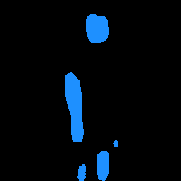}&
		\includegraphics[width=0.13\textwidth,height=1.9cm]
        {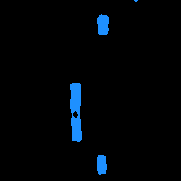}&
        \includegraphics[width=0.13\textwidth,height=1.9cm]
        {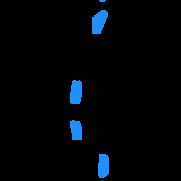}&
        \includegraphics[width=0.13\textwidth,height=1.9cm]
        {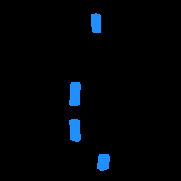}&
        \includegraphics[width=0.13\textwidth,height=1.9cm]
        {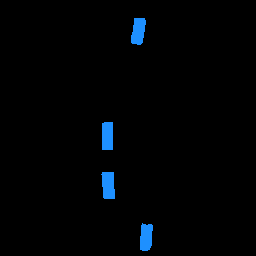}&
		\includegraphics[width=0.13\textwidth,height=1.9cm]
        {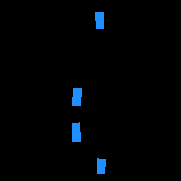}\\
		\includegraphics[width=0.22\textwidth,height=1.9cm]
        {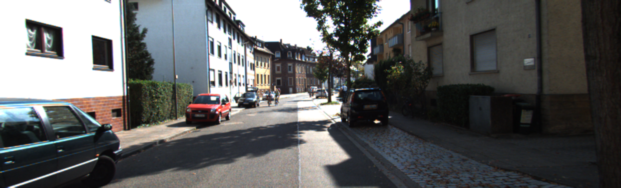}&
		\includegraphics[width=0.13\textwidth,height=1.9cm]
        {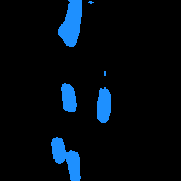}&
		\includegraphics[width=0.13\textwidth,height=1.9cm]
        {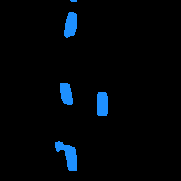}&
        \includegraphics[width=0.13\textwidth,height=1.9cm]
        {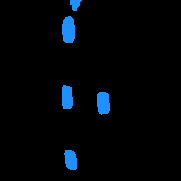}&
        \includegraphics[width=0.13\textwidth,height=1.9cm]
        {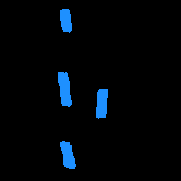}&
        \includegraphics[width=0.13\textwidth,height=1.9cm]
        {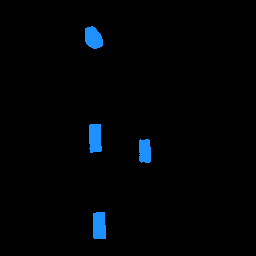}&
        \includegraphics[width=0.13\textwidth,height=1.9cm]
        {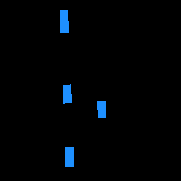}\\
        \includegraphics[width=0.22\textwidth,height=1.9cm]
        {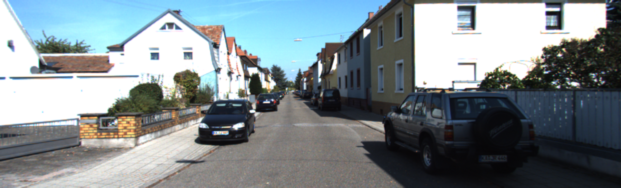}&
		\includegraphics[width=0.13\textwidth,height=1.9cm]
        {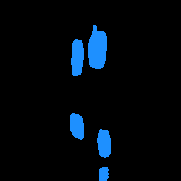}&
		\includegraphics[width=0.13\textwidth,height=1.9cm]
        {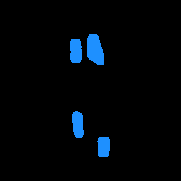}&
        \includegraphics[width=0.13\textwidth,height=1.9cm]
        {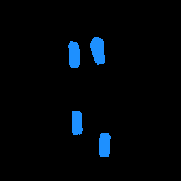}&
        \includegraphics[width=0.13\textwidth,height=1.9cm]
        {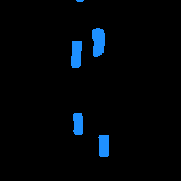}&
        \includegraphics[width=0.13\textwidth,height=1.9cm]
        {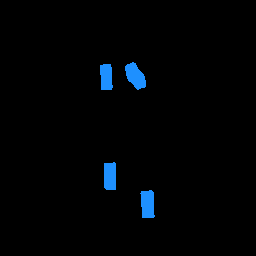}&
		\includegraphics[width=0.13\textwidth,height=1.9cm]
        {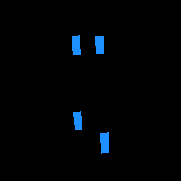}\\
		Front-view & Baseline & + MLP & + Correlation & + Cycle structure & + Feature selection & Ground-truth
	\end{tabular}
	\caption{Exemplar results of the ablation study for the front-to-top view projection module.}
	\label{fig:abla1}
\end{figure*}

\wx{In Fig.~\ref{fig:abla1}, we show several examples of the ablation study on the FTVP module. The examples are selected from the results of \textit{KITTI 3D Object} dataset. With the addition of structures (e.g., MLP, cross-view correlation, cycle structure, and feature selection) in the FTVP module, our model can effectively extract the masks of the individual vehicles, remove noises, and refine their shapes. }

\begin{table}[]
    \centering
    \setlength{\tabcolsep}{12pt}
    \caption{Effectiveness of the front-to-top view projection module.}
    \begin{tabular}{c|cc}
        \toprule
        Structure & mIOU (\%) & mAP (\%) \\
         \midrule
        Baseline & 22.31 & 34.58\\
        + MLP & 27.42 & 37.44\\
        + Cross-view Correlation & 35.03 & 46.33\\
        + Cycle Structure & 35.54 & 47.29 \\ 
        + Feature Selection & 39.97 & 54.53 \\
        + Multi-scale FTVPs & 37.83 & 55.02 \\
        + Deep Supervision & 40.69 & 59.05 \\
        \bottomrule
    \end{tabular}
    \label{tab:cvt1}
\end{table}

\begin{table}[]
    \centering
    \setlength{\tabcolsep}{12pt}
    \caption{Different input combinations of the cross-view transformer evaluated in \textit{KITTI 3D Object}.}
    \begin{tabular}{cc|cc}
        \toprule
         $K$ & $V$ & mIOU (\%) & mAP (\%) \\
         \midrule
        $X'$ & $X'$ & 39.19 & 53.20\\
        $X''$ & $X''$ & 40.12 & 53.49\\
        $X$ & $X$ & 39.55 & 56.67\\
        $X''$ & $X$ & 39.48 & 52.73\\
        $X$ & $X''$ & 40.69 & 59.05\\
        \bottomrule
    \end{tabular}
    \label{tab:cvt2}
\end{table}

\begin{figure}
    \centering
    \footnotesize
	\begin{tabular}{c@{}c@{}c}
        \includegraphics[width=0.19\textwidth,height=2.5cm]
        {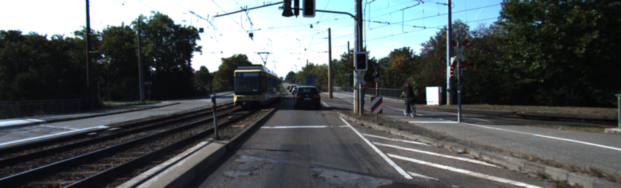} &
		\includegraphics[width=0.14\textwidth,height=2.5cm]
        {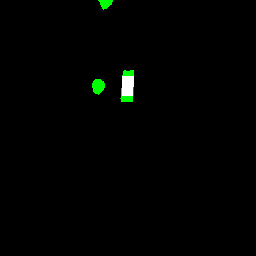} &
		\includegraphics[width=0.14\textwidth,height=2.5cm]
        {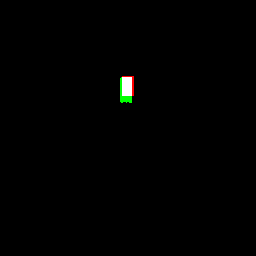} \\
        
        \includegraphics[width=0.19\textwidth,height=2.5cm]
        {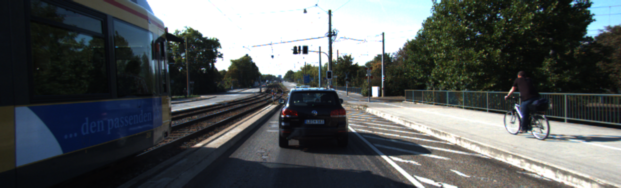} &
		\includegraphics[width=0.14\textwidth,height=2.5cm]
        {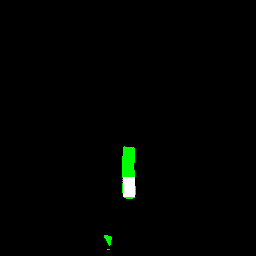} &
		\includegraphics[width=0.14\textwidth,height=2.5cm]
        {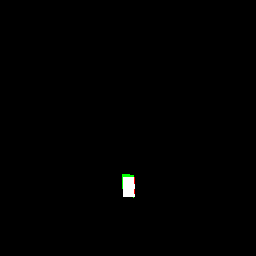} \\
        
        \includegraphics[width=0.19\textwidth,height=2.5cm]
        {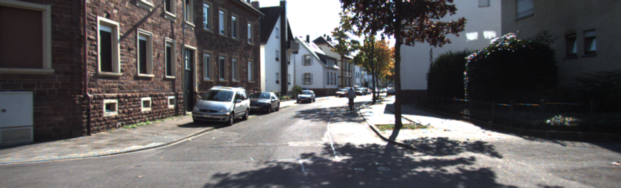} &
		\includegraphics[width=0.14\textwidth,height=2.5cm]
        {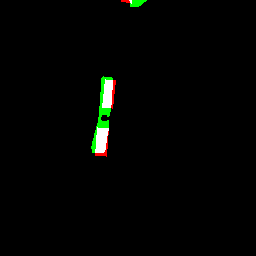} &
		\includegraphics[width=0.14\textwidth,height=2.5cm]
        {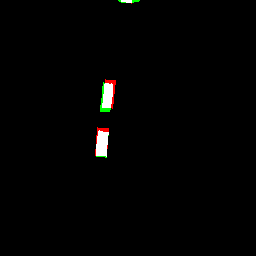} \\
        
        \includegraphics[width=0.19\textwidth,height=2.5cm]
        {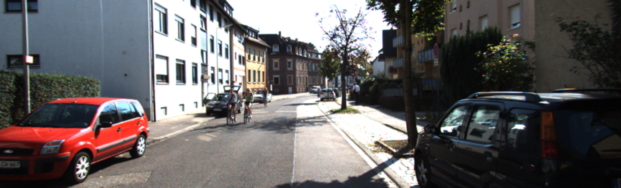} &
		\includegraphics[width=0.14\textwidth,height=2.5cm]
        {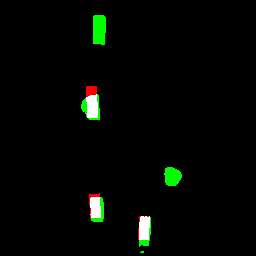} &
		\includegraphics[width=0.14\textwidth,height=2.5cm]
        {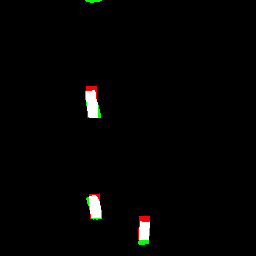} \\
        
        \includegraphics[width=0.19\textwidth,height=2.5cm]
        {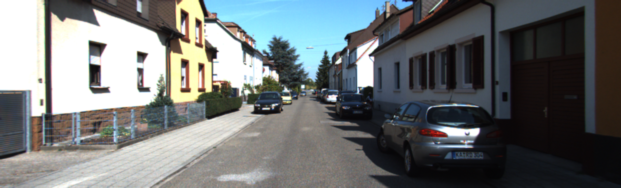} &
		\includegraphics[width=0.14\textwidth,height=2.5cm]
        {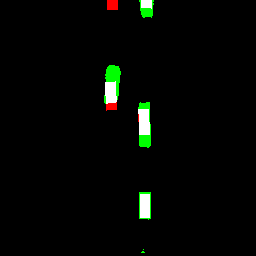} &
		\includegraphics[width=0.14\textwidth,height=2.5cm]
        {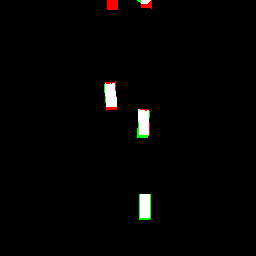} \\
        
        \includegraphics[width=0.19\textwidth,height=2.5cm]
        {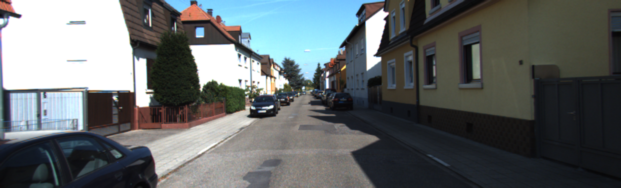} &
		\includegraphics[width=0.14\textwidth,height=2.5cm]
        {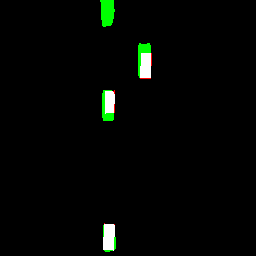} &
		\includegraphics[width=0.14\textwidth,height=2.5cm]
        {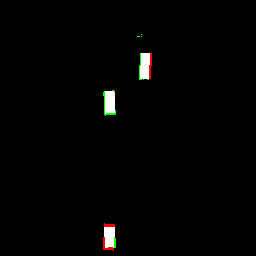} \\
        
        \includegraphics[width=0.19\textwidth,height=2.5cm]
        {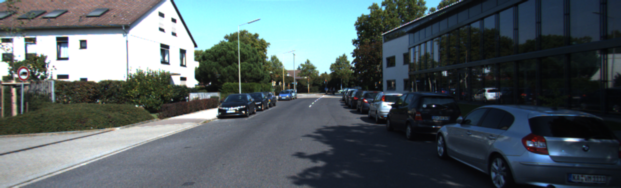} &
		\includegraphics[width=0.14\textwidth,height=2.5cm]
        {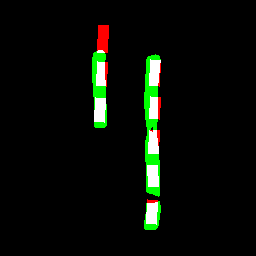} &
		\includegraphics[width=0.14\textwidth,height=2.5cm]
        {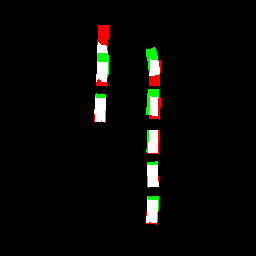} \\
        
        \includegraphics[width=0.19\textwidth,height=2.5cm]
        {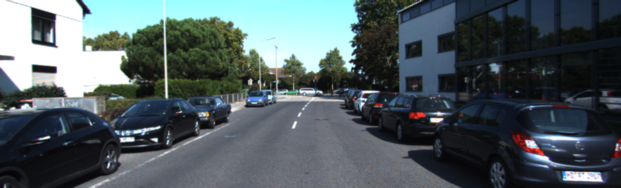} &
		\includegraphics[width=0.14\textwidth,height=2.5cm]
        {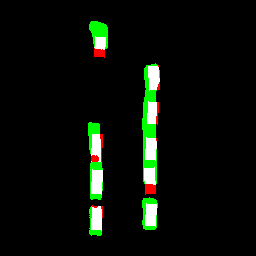} &
		\includegraphics[width=0.14\textwidth,height=2.5cm]
        {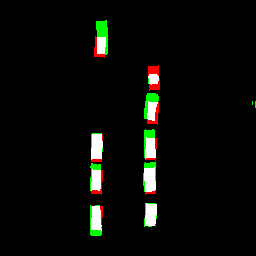} \\
		Front-view Image  & w/o FTVPs & w FTVPs
	\end{tabular}
	\caption{We illustrate that, while our model will cause spatial deviation with a single FTVP module, the proposed the multi-scale FTVP modules can propagate the rich spatial information to alleviate this problem. In addition, we highlight the pixels that are discrepancies between our estimation and the ground truth. The red and green pixels represent False Positives and False Negatives, respectively. Please zoom in for a better view.}
	\label{fig:skip}
\end{figure}


\textbf{Cross-view transformer.} 
We validate different input combinations of $K, Q, V$ for CVT. We demonstrate the results in Table \ref{tab:cvt2}. For all test cases, the query (i.e., $Q$) is assigned to a feature after view projection $X'$. As the most trivial case, we use $X'$ as $K$ and $V$ of CVT as well, which self-correlates all the non-local patches of $X'$. Since $X'$ may lose some information via view projection, CVT does not perform well. Because both $K$ and $V$ are assigned to $X$ or $X''$, it involves the features before view projection, but $X$ contains richer information than $X''$, which leads to better performance. Moreover, with $X$ and $X''$ corresponding to $K$ and $V$, the substantial information for view projection is implicitly introduced by $X''$ to strengthen the model. More specifically, using $X$ as the key is better for generating a precise relevance embedding, while applying $X''$ as the value encourages the involvement of most relevant features, leading to the optimal results. 

\textbf{Multi-scale FTVP modules.} Based on the model with a single FTVP module, we upgrade the network by integrating the multi-scale FTVP modules.
As observed in Table~\ref{tab:cvt1}, the upgraded model improves mAP while decreasing mIOU. This is perhaps because the features of different scales deliver rich, detailed information but their learning lacks guidance. To address this, we employ the deep supervision scheme. As shown, our model achieves the optimal performance (40.69\% mIOU and 59.07\% mAP). Note that, 
for reference, we also add deep supervision for the model with a single FTVP module, and its results are almost unchanged (39.48\% vs 39.97\% on mIOU and 54.78\% vs 54.53\% on mAP). Moreover, we show the representative visualization results with respect to the multi-scale FTVP modules in Fig.~\ref{fig:skip}. It is observed that the spatial deviation effect will be reduced in the presence of FTVPs.

\wx{To show the advantage of employing three FTVPs only, we compare our model against other variants, including FTVPs with local window attention, variants in which FTVPs have been replaced with naive convolutional layers, and variants with more than three FTVPs. The comparison results in terms of mIOU and mAP are demonstrated in Table~\ref{tab:cvt3}. However, local window FTVPs and convolutional layers are both local operations without the involvement of global information, while FTVP modules deployed on the shallow layers lack semantic information and demand a large number of parameters and a lot of computation for higher-resolution features. Thus, these variants all lead to sub-optimal performance, so we choose to retain three FTVPs in our network.}

\begin{table}[]
    \centering
    \setlength{\tabcolsep}{12pt}
    \caption{Advantage of retaining 3 FTVPs evaluated in \textit{KITTI 3D Object}. "-" represents out of memory.}
    \begin{tabular}{c|cc}
        \toprule
        Structure & mIOU (\%) & mAP (\%) \\
         \midrule
        Naive Conv  & 39.99 & 58.40 \\
        Window FTVPs  & 39.09 & 57.37 \\
        4 FTVPs & 39.69 & 57.70 \\
        5 or 6 FTVPs & - & - \\
        Ours (3 FTVPs) & 40.69 & 59.05 \\
        \bottomrule
    \end{tabular}
    \label{tab:cvt3}
\end{table}

\subsection{Network Efficiency}

\wxr{We measure FPS, the number of parameters, and FLOPs for the competing methods (i.e., VED \cite{lu2019monocular}, VPN \cite{pan2020cross}, PON \cite{roddick2020predicting}, and Stitch \cite{can2022understanding}) in Table~\ref{tab:fps}. All methods are tested on the same platform using a single NVIDIA Titan XP GPU. As observed, our model achieves real-time performance with few parameters. Thus, our model efficiency is comparable to these methods without using any model compression techniques.}

\subsection{Panorama HD Map Generation}

\begin{figure}
    \centering
    \includegraphics[width=0.5\textwidth,height=6cm]{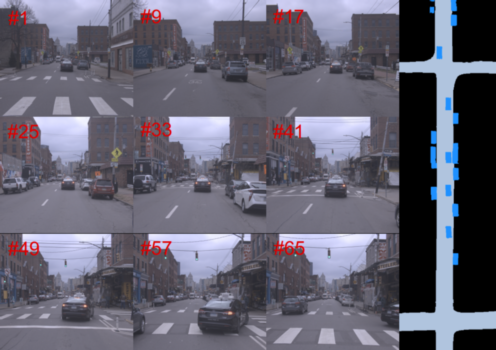}
    \includegraphics[width=0.5\textwidth,height=6cm]{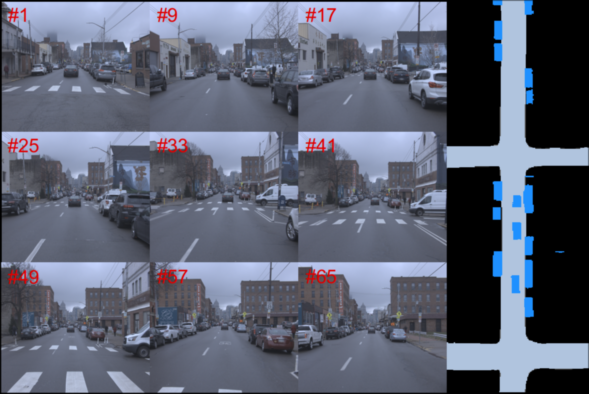}
    \includegraphics[width=0.5\textwidth,height=4.5cm]{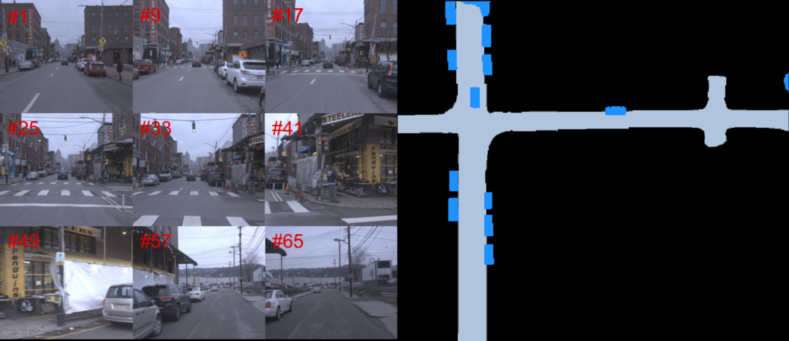}
    \caption{We montage the estimated road layout from the image sequences of \textit{Argoverse} to produce three panorama HD maps (on the right side of the figure) containing the road layout and vehicle occupancies. }
    \label{fig:hdmap}
\end{figure}

We showcase the application of our model on the \textit{Argoverse} dataset for generating a panorama HD map via stitching the road layout estimation given the consecutive front-view images. The generated HD map is shown in Fig. \ref{fig:hdmap}, highlighting the potential of our approach for generating panorama HD maps.

\subsection{Failure cases}

\begin{figure}
    \centering
	\footnotesize
	\begin{tabular}{c@{}c@{}c}
        \includegraphics[width=0.18\textwidth,height=2.4cm]
        {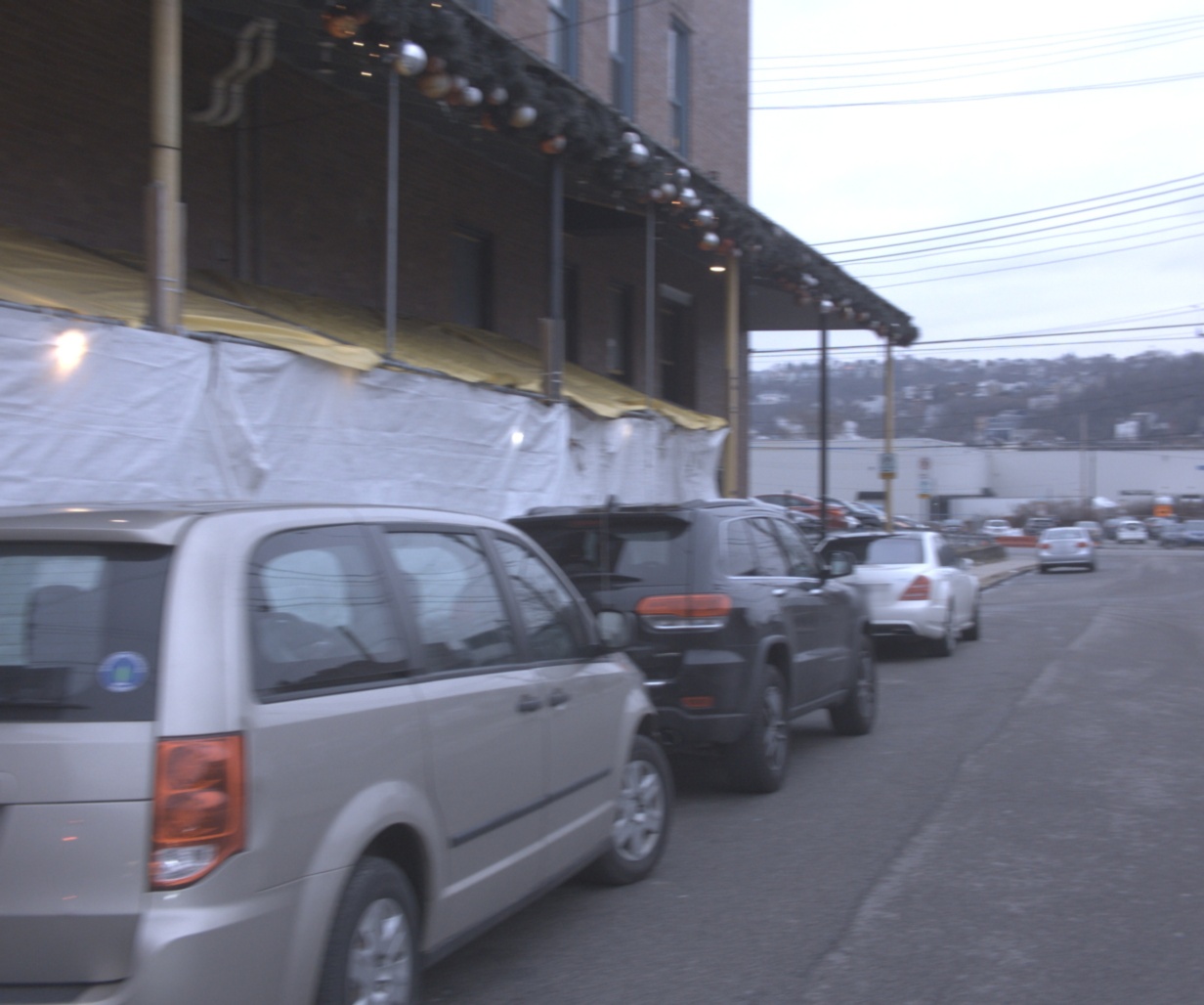} &
        \includegraphics[width=0.15\textwidth,height=2.4cm]
        {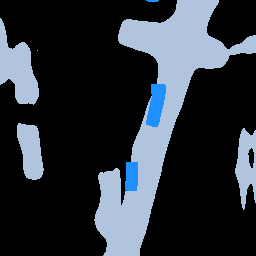}&
		\includegraphics[width=0.15\textwidth,height=2.4cm]
        {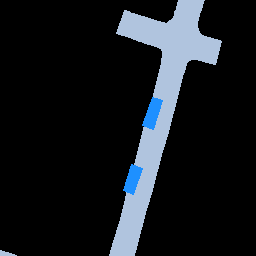}\\
        \includegraphics[width=0.18\textwidth,height=2.4cm]
        {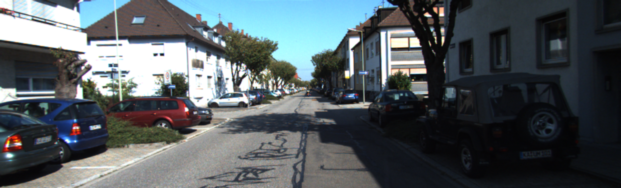} &
        \includegraphics[width=0.15\textwidth,height=2.4cm]
        {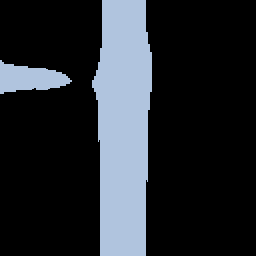}&
		\includegraphics[width=0.15\textwidth,height=2.4cm]
        {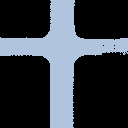}\\
		\includegraphics[width=0.18\textwidth,height=2.4cm]
        {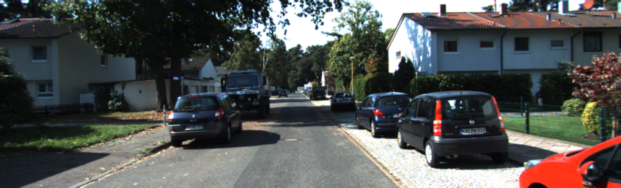} &
		\includegraphics[width=0.15\textwidth,height=2.4cm]
        {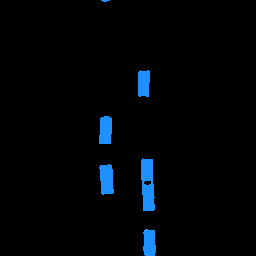}&
		\includegraphics[width=0.15\textwidth,height=2.4cm]
        {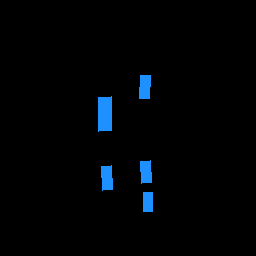}\\
        
		Front-View  & Ours & Ground-truth
	\end{tabular}
	\caption{Failure cases on \textit{Argoverse}, \textit{KITTI Odometry}, and \textit{KITTI 3D Object}.}
	\label{fig:fail}
\end{figure}

Fig.~\ref{fig:fail} shows a few scenarios in which we cannot produce an accurate road layout estimation. First, our model may produce inaccurate prediction in the cases of occlusion and sharp turns (see the first and second rows in Fig.~\ref{fig:fail}). Second, our model may be confused by multiple close vehicles (see the third row in Fig.~\ref{fig:fail}).

\begin{table}[t]
    \centering
    \footnotesize
    \setlength{\tabcolsep}{20pt}
    \caption{Comparison of model efficiency.}
    \begin{tabular}{c|c|c|c}
        \toprule
        Model & {FPS} & {Params} & {FLOPs}  \\ 
        \midrule\midrule
        VED \cite{lu2019monocular} & 77 & 45.61M & 140.58G \\
        VPN \cite{pan2020cross} & 100 & 18.29M & 37.61G \\
        PON \cite{roddick2020predicting} & 17 & 38.64M & 96.12G \\
        Stitch \cite{can2022understanding} & 6 & 52.32M & 589.25G \\
        Ours & 25 & 24.43M & 48.04G \\
        \bottomrule
    \end{tabular}
    \label{tab:fps}
\end{table}

\section{Conclusion}
\label{sec:Conclusion}
In this paper, we present a novel framework to estimate road layout and vehicle occupancy in top-views given a front-view monocular image. We propose a front-to-top view projection module that is composed of a cycled view projection structure and a cross-view transformer in which the features of the views before and after projection are explicitly correlated and the most relevant features for view projection are fully exploited in order to enhance the transformed features. 
\wx{In addition, we introduce the multi-scale FTVP modules to compensate for the information loss in the encoded features. We also demonstrate that our proposed model achieves the state-of-the-art performance and runs at $25$ FPS on a single GPU, which is efficient and applicable for real-time panorama HD map reconstruction.}

\wx{Nevertheless, the images or videos captured from real life for the vision tasks are bound to cause class-imbalance problems. Especially for our task, the sample ratio between some classes reaches 100-1000 times. Thus, it remains very challenging for our model to predict those classes with small samples, which leads to sub-optimal performance in certain classes for our results. In addition, errors may occur when vehicles are very close to each other, encounter the occlusions, or make sharp turns due to the large view gap.}

\wx{For our task, the major issues rest in the class imbalance and large view gap. As the future work, we will consider to exploit synthetic data to supplement the small sample classes for data balance to strengthen our model performance. On the other hand, we will further exploit the information from the front-view images and explore more effective transformation between different views.}



%





\ifCLASSOPTIONcaptionsoff
  \newpage
\fi



%


\bibliography{egbib}{}

\begin{thebibliography}{10}
\providecommand{\url}[1]{#1}
\csname url@samestyle\endcsname
\providecommand{\newblock}{\relax}
\providecommand{\bibinfo}[2]{#2}
\providecommand{\BIBentrySTDinterwordspacing}{\spaceskip=0pt\relax}
\providecommand{\BIBentryALTinterwordstretchfactor}{4}
\providecommand{\BIBentryALTinterwordspacing}{\spaceskip=\fontdimen2\font plus
\BIBentryALTinterwordstretchfactor\fontdimen3\font minus
  \fontdimen4\font\relax}
\providecommand{\BIBforeignlanguage}[2]{{%
\expandafter\ifx\csname l@#1\endcsname\relax
\typeout{** WARNING: IEEEtran.bst: No hyphenation pattern has been}%
\typeout{** loaded for the language `#1'. Using the pattern for}%
\typeout{** the default language instead.}%
\else
\language=\csname l@#1\endcsname
\fi
#2}}
\providecommand{\BIBdecl}{\relax}
\BIBdecl

\bibitem{Geiger14}
A.~{Geiger}, M.~{Lauer}, C.~{Wojek}, C.~{Stiller}, and R.~{Urtasun}, ``3d
  traffic scene understanding from movable platforms,'' \emph{PAMI}, vol.~36,
  no.~5, pp. 1012--1025, 2014.

\bibitem{sun2019leveraging}
T.~Sun, Z.~Di, P.~Che, C.~Liu, and Y.~Wang, ``Leveraging crowdsourced gps data
  for road extraction from aerial imagery,'' in \emph{CVPR}, 2019, pp.
  7509--7518.

\bibitem{roddick2020predicting}
T.~Roddick and R.~Cipolla, ``Predicting semantic map representations from
  images using pyramid occupancy networks,'' in \emph{CVPR}, 2020, pp.
  11\,138--11\,147.

\bibitem{liu2020understanding}
B.~Liu, B.~Zhuang, S.~Schulter, P.~Ji, and M.~Chandraker, ``Understanding road
  layout from videos as a whole,'' in \emph{CVPR}, 2020, pp. 4414--4423.

\bibitem{liang2019convolutional}
J.~Liang, N.~Homayounfar, W.-C. Ma, S.~Wang, and R.~Urtasun, ``Convolutional
  recurrent network for road boundary extraction,'' in \emph{CVPR}, 2019, pp.
  9512--9521.

\bibitem{2018Learning}
S.~Schulter, M.~Zhai, N.~Jacobs, and M.~Chandraker, ``Learning to look around
  objects for top-view representations of outdoor scenes,'' in \emph{ECCV},
  2018.

\bibitem{2019A}
Z.~Wang, B.~Liu, S.~Schulter, and M.~Chandraker, ``A parametric top-view
  representation of complex road scenes,'' in \emph{CVPR}, 2019.

\bibitem{2016Monocular}
X.~Chen, K.~Kundu, Z.~Zhang, H.~Ma, and R.~Urtasun, ``Monocular 3d object
  detection for autonomous driving,'' in \emph{CVPR}, 2016.

\bibitem{srivastava2019learning}
S.~Srivastava, F.~Jurie, and G.~Sharma, ``Learning 2d to 3d lifting for object
  detection in 3d for autonomous vehicles,'' in \emph{2019 IEEE/RSJ
  International Conference on Intelligent Robots and Systems (IROS)}.\hskip 1em
  plus 0.5em minus 0.4em\relax IEEE, 2019, pp. 4504--4511.

\bibitem{roddick2018orthographic}
T.~Roddick, A.~Kendall, and R.~Cipolla, ``Orthographic feature transform for
  monocular 3d object detection,'' in \emph{BMVC}, 2019.

\bibitem{simonelli2019disentangling}
A.~Simonelli, S.~R. Bulo, L.~Porzi, M.~L{\'o}pez-Antequera, and
  P.~Kontschieder, ``Disentangling monocular 3d object detection,'' in
  \emph{ICCV}, 2019, pp. 1991--1999.

\bibitem{ding2020learning}
M.~Ding, Y.~Huo, H.~Yi, Z.~Wang, J.~Shi, Z.~Lu, and P.~Luo, ``Learning
  depth-guided convolutions for monocular 3d object detection,'' in
  \emph{CVPRW}, 2020, pp. 1000--1001.

\bibitem{2020GS3D}
B.~Li, W.~Ouyang, L.~Sheng, X.~Zeng, and X.~Wang, ``Gs3d: An efficient 3d
  object detection framework for autonomous driving,'' in \emph{CVPR}, 2020.

\bibitem{mozaffari2020deep}
S.~Mozaffari, O.~Y. Al-Jarrah, M.~Dianati, P.~Jennings, and A.~Mouzakitis,
  ``Deep learning-based vehicle behavior prediction for autonomous driving
  applications: A review,'' \emph{IEEE Transactions on Intelligent
  Transportation Systems}, 2020.

\bibitem{hong2019rules}
J.~Hong, B.~Sapp, and J.~Philbin, ``Rules of the road: Predicting driving
  behavior with a convolutional model of semantic interactions,'' in
  \emph{CVPR}, 2019, pp. 8454--8462.

\bibitem{kim2020advisable}
J.~Kim, S.~Moon, A.~Rohrbach, T.~Darrell, and J.~Canny, ``Advisable learning
  for self-driving vehicles by internalizing observation-to-action rules,'' in
  \emph{CVPR}, 2020, pp. 9661--9670.

\bibitem{ma2019trafficpredict}
Y.~Ma, X.~Zhu, S.~Zhang, R.~Yang, W.~Wang, and D.~Manocha, ``Trafficpredict:
  Trajectory prediction for heterogeneous traffic-agents,'' in \emph{AAAI},
  vol.~33, 2019, pp. 6120--6127.

\bibitem{hou2019learning}
Y.~Hou, Z.~Ma, C.~Liu, and C.~C. Loy, ``Learning lightweight lane detection
  cnns by self attention distillation,'' in \emph{ICCV}, 2019, pp. 1013--1021.

\bibitem{zou2019robust}
Q.~Zou, H.~Jiang, Q.~Dai, Y.~Yue, L.~Chen, and Q.~Wang, ``Robust lane detection
  from continuous driving scenes using deep neural networks,'' \emph{IEEE
  transactions on vehicular technology}, vol.~69, no.~1, pp. 41--54, 2019.

\bibitem{philion2019fastdraw}
J.~Philion, ``Fastdraw: Addressing the long tail of lane detection by adapting
  a sequential prediction network,'' in \emph{CVPR}, 2019, pp.
  11\,582--11\,591.

\bibitem{lin2012vision}
C.-C. Lin and M.-S. Wang, ``A vision based top-view transformation model for a
  vehicle parking assistant,'' \emph{Sensors}, vol.~12, no.~4, pp. 4431--4446,
  2012.

\bibitem{tseng2013image}
D.~C. Tseng, T.~W. Chao, and J.~W. Chang, ``Image-based parking guiding using
  ackermann steering geometry,'' in \emph{Applied Mechanics and Materials},
  vol. 437.\hskip 1em plus 0.5em minus 0.4em\relax Trans Tech Publ, 2013, pp.
  823--826.

\bibitem{zhu2018generative}
X.~Zhu, Z.~Yin, J.~Shi, H.~Li, and D.~Lin, ``Generative adversarial frontal
  view to bird view synthesis,'' in \emph{2018 International conference on 3D
  Vision (3DV)}.\hskip 1em plus 0.5em minus 0.4em\relax IEEE, 2018, pp.
  454--463.

\bibitem{regmi2018cross}
K.~Regmi and A.~Borji, ``Cross-view image synthesis using conditional gans,''
  in \emph{CVPR}, 2018, pp. 3501--3510.

\bibitem{yang2021projecting}
W.~Yang, Q.~Li, W.~Liu, Y.~Yu, Y.~Ma, S.~He, and J.~Pan, ``Projecting your view
  attentively: Monocular road scene layout estimation via cross-view
  transformation,'' in \emph{CVPR}, 2021, pp. 15\,536--15\,545.

\bibitem{fan2020sne}
R.~Fan, H.~Wang, P.~Cai, and M.~Liu, ``Sne-roadseg: Incorporating surface
  normal information into semantic segmentation for accurate freespace
  detection,'' in \emph{ECCV}.\hskip 1em plus 0.5em minus 0.4em\relax Springer,
  2020, pp. 340--356.

\bibitem{teichmann2018multinet}
M.~Teichmann, M.~Weber, M.~Zoellner, R.~Cipolla, and R.~Urtasun, ``Multinet:
  Real-time joint semantic reasoning for autonomous driving,'' in \emph{2018
  IEEE Intelligent Vehicles Symposium (IV)}.\hskip 1em plus 0.5em minus
  0.4em\relax IEEE, 2018, pp. 1013--1020.

\bibitem{yang2018denseaspp}
M.~Yang, K.~Yu, C.~Zhang, Z.~Li, and K.~Yang, ``Denseaspp for semantic
  segmentation in street scenes,'' in \emph{CVPR}, 2018, pp. 3684--3692.

\bibitem{yu2018bisenet}
C.~Yu, J.~Wang, C.~Peng, C.~Gao, G.~Yu, and N.~Sang, ``Bisenet: Bilateral
  segmentation network for real-time semantic segmentation,'' in \emph{ECCV},
  2018, pp. 325--341.

\bibitem{fu2019dual}
J.~Fu, J.~Liu, H.~Tian, Y.~Li, Y.~Bao, Z.~Fang, and H.~Lu, ``Dual attention
  network for scene segmentation,'' in \emph{CVPR}, 2019, pp. 3146--3154.

\bibitem{2012Automatic}
S.~Sengupta, P.~Sturgess, L.~Ladicky, and P.~H. Torr, ``Automatic dense visual
  semantic mapping from street-level imagery,'' in \emph{IEEE/RSJ International
  Conference on Intelligent Robots \& Systems}, 2012.

\bibitem{2016HD}
G.~Mattyus, S.~Wang, S.~Fidler, and R.~Urtasun, ``Hd maps: Fine-grained road
  segmentation by parsing ground and aerial images,'' in \emph{CVPR}, 2016.

\bibitem{2017Predicting}
M.~Zhai, Z.~Bessinger, S.~Workman, and N.~Jacobs, ``Predicting ground-level
  scene layout from aerial imagery,'' in \emph{CVPR}, 2017.

\bibitem{2017Cognitive}
S.~Gupta, V.~Tolani, J.~Davidson, S.~Levine, R.~Sukthankar, and J.~Malik,
  ``Cognitive mapping and planning for visual navigation,'' \emph{IJCV}, no.~4,
  2017.

\bibitem{lu2019monocular}
C.~Lu, M.~J.~G. van~de Molengraft, and G.~Dubbelman, ``Monocular semantic
  occupancy grid mapping with convolutional variational encoder--decoder
  networks,'' \emph{IEEE Robotics and Automation Letters}, vol.~4, no.~2, pp.
  445--452, 2019.

\bibitem{pan2020cross}
B.~Pan, J.~Sun, H.~Y.~T. Leung, A.~Andonian, and B.~Zhou, ``Cross-view semantic
  segmentation for sensing surroundings,'' \emph{IEEE Robotics and Automation
  Letters}, vol.~5, no.~3, pp. 4867--4873, 2020.

\bibitem{philion2020lift}
J.~Philion and S.~Fidler, ``Lift, splat, shoot: Encoding images from arbitrary
  camera rigs by implicitly unprojecting to 3d,'' in \emph{ECCV}.\hskip 1em
  plus 0.5em minus 0.4em\relax Springer, 2020, pp. 194--210.

\bibitem{2017Mousavian}
A.~Mousavian, D.~Anguelov, J.~Flynn, and J.~Kosecka, ``3d bounding box
  estimation using deep learning and geometry,'' in \emph{CVPR}, 2017.

\bibitem{lang2019pointpillars}
A.~H. Lang, S.~Vora, H.~Caesar, L.~Zhou, J.~Yang, and O.~Beijbom,
  ``Pointpillars: Fast encoders for object detection from point clouds,'' in
  \emph{CVPR}, 2019, pp. 12\,697--12\,705.

\bibitem{dwivedi2021bird}
I.~Dwivedi, S.~Malla, Y.-T. Chen, and B.~Dariush, ``Bird’s eye view
  segmentation using lifted 2d semantic features,'' 2021.

\bibitem{saha2021enabling}
A.~Saha, O.~Mendez, C.~Russell, and R.~Bowden, ``Enabling spatio-temporal
  aggregation in birds-eye-view vehicle estimation,'' in \emph{ICRA}.\hskip 1em
  plus 0.5em minus 0.4em\relax IEEE, 2021, pp. 5133--5139.

\bibitem{can2022understanding}
Y.~B. Can, A.~Liniger, O.~Unal, D.~Paudel, and L.~Van~Gool, ``Understanding
  bird’s-eye view of road semantics using an onboard camera,'' \emph{IEEE
  Robotics and Automation Letters}, vol.~7, no.~2, pp. 3302--3309, 2022.

\bibitem{wang2019monocular}
D.~{Wang}, C.~{Devin}, Q.~{Cai}, P.~{Krähenbühl}, and T.~{Darrell},
  ``Monocular plan view networks for autonomous driving,'' in \emph{2019
  IEEE/RSJ International Conference on Intelligent Robots and Systems (IROS)},
  2019, pp. 2876--2883.

\bibitem{2020MonoLayout}
K.~Mani, S.~Daga, S.~Garg, N.~S. Shankar, K.~M. Jatavallabhula, and K.~M.
  Krishna, ``Monolayout: Amodal scene layout from a single image,'' in
  \emph{WACV}, 2020.

\bibitem{huang2018lane}
Y.~Huang, Y.~Li, X.~Hu, and W.~Ci, ``Lane detection based on inverse
  perspective transformation and kalman filter.'' \emph{KSII Transactions on
  Internet \& Information Systems}, vol.~12, no.~2, 2018.

\bibitem{abbas2019geometric}
S.~A. Abbas and A.~Zisserman, ``A geometric approach to obtain a bird's eye
  view from an image.'' in \emph{ICCV Workshops}, 2019, pp. 4095--4104.

\bibitem{tang2019multi}
H.~Tang, D.~Xu, N.~Sebe, Y.~Wang, J.~J. Corso, and Y.~Yan, ``Multi-channel
  attention selection gan with cascaded semantic guidance for cross-view image
  translation,'' in \emph{CVPR}, 2019, pp. 2417--2426.

\bibitem{vaswani2017attention}
A.~Vaswani, N.~Shazeer, N.~Parmar, J.~Uszkoreit, L.~Jones, A.~N. Gomez,
  {\L}.~Kaiser, and I.~Polosukhin, ``Attention is all you need,'' in
  \emph{Advances in neural information processing systems}, 2017, pp.
  5998--6008.

\bibitem{zhang2020feature}
D.~Zhang, H.~Zhang, J.~Tang, M.~Wang, X.~Hua, and Q.~Sun, ``Feature pyramid
  transformer,'' in \emph{ECCV}, 2020.

\bibitem{carion2020end}
N.~Carion, F.~Massa, G.~Synnaeve, N.~Usunier, A.~Kirillov, and S.~Zagoruyko,
  ``End-to-end object detection with transformers,'' in \emph{ECCV}, 2020.

\bibitem{zhu2020deformable}
X.~Zhu, W.~Su, L.~Lu, B.~Li, X.~Wang, and J.~Dai, ``Deformable detr: Deformable
  transformers for end-to-end object detection,'' \emph{arXiv preprint
  arXiv:2010.04159}, 2020.

\bibitem{gavrilyuk2020actor}
K.~Gavrilyuk, R.~Sanford, M.~Javan, and C.~G. Snoek, ``Actor-transformers for
  group activity recognition,'' in \emph{CVPR}, 2020, pp. 839--848.

\bibitem{yang2020learning}
F.~Yang, H.~Yang, J.~Fu, H.~Lu, and B.~Guo, ``Learning texture transformer
  network for image super-resolution,'' in \emph{CVPR}, 2020, pp. 5791--5800.

\bibitem{dosovitskiy2021an}
A.~Dosovitskiy, L.~Beyer, A.~Kolesnikov, D.~Weissenborn, X.~Zhai,
  T.~Unterthiner, M.~Dehghani, M.~Minderer, G.~Heigold, S.~Gelly, J.~Uszkoreit,
  and N.~Houlsby, ``An image is worth 16x16 words: Transformers for image
  recognition at scale,'' in \emph{ICLR}, 2021.

\bibitem{han2021transformer}
K.~Han, A.~Xiao, E.~Wu, J.~Guo, C.~Xu, and Y.~Wang, ``Transformer in
  transformer,'' \emph{NIPS}, vol.~34, pp. 15\,908--15\,919, 2021.

\bibitem{liu2021swin}
Z.~Liu, Y.~Lin, Y.~Cao, H.~Hu, Y.~Wei, Z.~Zhang, S.~Lin, and B.~Guo, ``Swin
  transformer: Hierarchical vision transformer using shifted windows,'' in
  \emph{ICCV}, 2021, pp. 10\,012--10\,022.

\bibitem{wang2021pyramid}
W.~Wang, E.~Xie, X.~Li, D.-P. Fan, K.~Song, D.~Liang, T.~Lu, P.~Luo, and
  L.~Shao, ``Pyramid vision transformer: A versatile backbone for dense
  prediction without convolutions,'' in \emph{ICCV}, 2021, pp. 568--578.

\bibitem{he2016deep}
K.~He, X.~Zhang, S.~Ren, and J.~Sun, ``Deep residual learning for image
  recognition,'' in \emph{CVPR}, 2016, pp. 770--778.

\bibitem{zhu2017unpaired}
J.-Y. Zhu, T.~Park, P.~Isola, and A.~A. Efros, ``Unpaired image-to-image
  translation using cycle-consistent adversarial networks,'' in \emph{ICCV},
  2017, pp. 2223--2232.

\bibitem{dwibedi2019temporal}
D.~Dwibedi, Y.~Aytar, J.~Tompson, P.~Sermanet, and A.~Zisserman, ``Temporal
  cycle-consistency learning,'' in \emph{CVPR}, 2019, pp. 1801--1810.

\bibitem{kingmaadam}
D.~P. Kingma and J.~L. Ba, ``Adam: Amethod for stochastic optimization.''

\bibitem{geiger2012we}
A.~Geiger, P.~Lenz, and R.~Urtasun, ``Are we ready for autonomous driving? the
  kitti vision benchmark suite,'' in \emph{CVPR}.\hskip 1em plus 0.5em minus
  0.4em\relax IEEE, 2012, pp. 3354--3361.

\bibitem{chang2019argoverse}
M.-F. Chang, J.~Lambert, P.~Sangkloy, J.~Singh, S.~Bak, A.~Hartnett, D.~Wang,
  P.~Carr, S.~Lucey, D.~Ramanan \emph{et~al.}, ``Argoverse: 3d tracking and
  forecasting with rich maps,'' in \emph{CVPR}, 2019, pp. 8748--8757.

\bibitem{caesar2020nuscenes}
H.~Caesar, V.~Bankiti, A.~H. Lang, S.~Vora, V.~E. Liong, Q.~Xu, A.~Krishnan,
  Y.~Pan, G.~Baldan, and O.~Beijbom, ``nuscenes: A multimodal dataset for
  autonomous driving,'' in \emph{CVPR}, 2020, pp. 11\,621--11\,631.

\bibitem{behley2019semantickitti}
J.~Behley, M.~Garbade, A.~Milioto, J.~Quenzel, S.~Behnke, C.~Stachniss, and
  J.~Gall, ``Semantickitti: A dataset for semantic scene understanding of lidar
  sequences,'' in \emph{ICCV}, 2019, pp. 9297--9307.

\end{thebibliography}
\bibliographystyle{IEEEtran}

%








\end{document}